\documentclass{article}

 \usepackage[final]{colm2026_conference}

\usepackage[utf8]{inputenc} 
\usepackage[T1]{fontenc}    
\usepackage{url}            
\usepackage{booktabs}       
\usepackage{amsfonts}       
\usepackage{nicefrac}       
\usepackage{microtype}      
\usepackage{xcolor}         
\usepackage[table]{xcolor}
\usepackage{lipsum}
\usepackage{pgfplotstable}

\usepackage{amsmath}
\usepackage{amssymb}
\usepackage{mathtools}
\usepackage{enumitem}

\usepackage{amsthm}

\usepackage{multirow}       

\usepackage{url}
\usepackage[utf8]{inputenc} 
\usepackage[T1]{fontenc}    
\usepackage[breaklinks]{hyperref}      

\usepackage{booktabs}       
\usepackage{amsfonts}       
\usepackage{amsmath}
\usepackage[mathscr]{euscript}
\usepackage{amssymb}
\usepackage{mathtools}
\usepackage{amsthm}
\usepackage{nicefrac}       
\usepackage{microtype}      

\usepackage{graphicx}
\usepackage{xcolor}      

\usepackage{caption}
\usepackage{multirow,makecell}

\usepackage{xspace}
\usepackage{subcaption}

\usepackage{algorithm}
\usepackage{algorithmic}
\usepackage{wrapfig}

\usepackage[most]{tcolorbox}

\usepackage{lineno}

\usepackage[capitalize,noabbrev]{cleveref}

\theoremstyle{plain}

\theoremstyle{definition}

\theoremstyle{remark}

\title{Reasoning Shift: How Context Silently Shortens LLM Reasoning}

\author{
Gleb Rodionov$^{\dagger,*}$ \And
Roman Garipov$^{\S, \diamond}$ \And
George Yakushev$^{\ddagger,\dagger}$
}

\begin{document}

\ifcolmsubmission
\linenumbers
\fi

\maketitle

\makeatletter
\def\@trackname{}
\makeatother
\maketitle

\begingroup
\makeatletter
\renewcommand\@makefnmark{}
\footnotetext{$^\dagger$ Yandex\
$^\ddagger$ HSE University\
$^\S$ New York University.\\
$\diamond$ - work done during employement at HSE University and Yandex Research.\\
Correspondence to: \texttt{rodionovgleb@yandex-team.ru}\,.}
\makeatother
\endgroup

\vspace{-8px}
\begin{abstract}
Large language models (LLMs) exhibiting test-time scaling behavior, such as extended reasoning traces and self-verification, have demonstrated remarkable performance on complex, long-term reasoning tasks. However, the robustness of these reasoning behaviors remains underexplored. 
To investigate this, we conduct a systematic evaluation of multiple reasoning models across three scenarios: (1) problems augmented with lengthy, irrelevant context; (2) multi-turn conversational settings with independent tasks; and (3) problems presented as a subtask within a complex task. We observe an interesting phenomenon: reasoning models tend to produce much shorter reasoning traces (up to 74\%) for the same problem under different context conditions compared to the traces produced when the problem is presented in isolation. A finer-grained analysis reveals that this compression is associated with a decrease in self-verification and uncertainty management behaviors, such as double-checking. Importantly, we show that even when additional self-checks are forced, their efficiency depends not only on the content of the reasoning traces, but also on the presence of redundant context. We hope our findings draw additional attention to both the robustness of reasoning models and the problem of context management for LLMs.
\end{abstract}

\vspace{-10px}\section{Introduction}\label{sect:intro}\vspace{-10px}

Recently, test-time scaling has emerged as a promising strategy for improving the performance of Large Language Models by allocating more compute during inference, allowing adaptation to input complexity without retraining \citep{openai_o1, deepseek_r1}. A prominent example of test-time scaling is Chain-of-Thought (CoT) \citep{zero_shot_cot_Kojima2022LargeLM, cot_wei_2022}. With recent advancements in Reinforcement Learning (RL) and other post-training methods, LLMs are now equipped with a thinking mode, which enables them to produce long reasoning trajectories before providing an answer to the user's query. Importantly, modern advancements in reasoning LLMs are often tied to the models' ability to self-reflect on how they solve problems, as demonstrated by the presence of high-level patterns in their CoT, such as backtracking, uncertainty management, and self-verification \citep{gandhi2025cognitive_habits, venhoff2025understanding}. These patterns typically accompany increased reasoning trace length.

In parallel, recent years have seen the rise of long-context language models, with context windows expanding to hundreds of thousands or even millions of tokens \citep{dubey2024llama, anthropic2024claude3, googledeepmind2025gemini25thinking}. These advances provide ample space for test-time scaling, allowing models to explore, reflect, and summarize within a single context, thereby enabling complex reasoning \citep{liu2025deepseekv3_2}, multi-stage prover-verifier pipelines \citep{shao2025deepseekmathv2}, and sophisticated agent workflows \citep{team2025kimik2}. However, multiple works have demonstrated significant limitations associated with longer contexts, such as reduced ability to retrieve relevant data from long contexts \citep{needleinhaystack}, to learn in context \citep{li2024long_icl_bench}, to recover from wrong assumptions in multi-turn conversations \citep{laban2026llms_get_lost}, and to perform multi-step reasoning over long inputs \citep{ling2025longreason}.

With the rise of multiple test-time strategies and agents working on long-term tasks, natural questions arise: "How do context length and content affect the reasoning capabilities of the models?" In particular, "If a model faces an isolated subproblem with irrelevant data in the context, will it solve it similarly to working on it in isolation?"

In this work, we study a surprising phenomenon: we observe a significant distribution shift in how models solve the same problems under different context conditions. We explore how reasoning quality and performance change under simple distracting conditions: (1) problems augmented with lengthy, irrelevant context; (2) multi-turn conversational settings with independent tasks; and (3) problems presented as subtasks within a complex task. We find that reasoning models tend to produce significantly fewer reasoning tokens when solving problems under non-isolated context conditions. Interestingly, we found that the part of the reasoning traces after finding the first candidate answer is disproportionately shortened compared to the time needed to find the first candidate answer. An analysis of the reasoning chains shows that this compression is associated with a decrease in self-verification and uncertainty management behaviors, such as double-checking. Importantly, we show that even when additional self-checks are forced, their efficiency depends not only on the content of the reasoning traces, but also on the presence of redundant context. Finally, we investigate several mitigation strategies and show that prompting and targeted supervised fine-tuning have limited effect on improving robustness to irrelevant context.

\vspace{-10px}
\section{Background}\label{sect:background}
\vspace{-10px}

\textbf{Chain-of-Thought reasoning.} Test-time scaling has brought a paradigm shift that enables long Chain-of-Thought reasoning and induces sophisticated reasoning behaviors, making models superior in competitive math and coding tasks. The central technique driving this revolution is large-scale RL, which elicits complex reasoning behaviors such as self-verification and iterative refinement \citep{openai_o1, deepseek_r1, shao2024deepseekmath, yu2025dapo, kumar2024training_to_self_correct}. Subsequent works have explored additional approaches to achieve test-time scaling and strong reasoning performance \citep{muennighoff2025s1, ye2025limoreasoning}.

Additional thinking budget introduces new challenges in adjusting reasoning effort according to input problem complexity \citep{qwen3}. \cite{su2025between_underthinking} analyzed the relationship between reasoning length and answer correctness, finding that LLMs tend to overthink on simpler problems and underthink on harder ones, indicating that models may fail to calibrate their reasoning length accordingly. \cite{aggarwal2025optimalthinkingbench} proposed a unified benchmark that jointly evaluates overthinking and underthinking in LLMs.

To analyze how LLMs tackle hard reasoning problems within their long CoT traces, \cite{venhoff2025understanding} define distinct reasoning functions within a reasoning trace and use an LLM judge to classify each sentence according to its functional role. Adopting their framework, \cite{bogdan2025thought_anchors} propose analyzing long reasoning traces by identifying steps that guide the trajectory of reasoning, organizing sentences into different categories and measuring their causal impact, highlighting the importance of planning or uncertainty management sentences. \cite{muennighoff2025s1} also demonstrated that enforcing self-reflection by intervening in the reasoning trace when the model attempts to stop may lead to improved performance. Subsequent works have investigated the relationship between accuracy and high-level characteristics of reasoning traces, such as length, review ratio, and others \citep{wu2025understand_CoT_lenth, hassid2025dont_overthink, feng2025characterizes_reasoning}.

\textbf{Long context and context management.} Recent years have witnessed remarkable growth in model context length \citep{dubey2024llama, anthropic2024claude3, googledeepmind2025gemini25thinking}. Despite this impressive scaling of context extension, a significant gap remains between the context length these models claim to support and the actual context length they can process effectively \citep{liu2025long_context_survey}. Common limitations typically involve retrieval-based evaluations \citep{needleinhaystack} and multi-hop tracing and aggregation \citep{hsieh2024ruler}. However, \cite{du2025context_length_alone} show that long contexts may degrade performance despite perfect retrieval, even in a synthetic setting where models are forced to attend only to relevant tokens. \cite{laban2026llms_get_lost} demonstrate significant performance drops in multi-turn settings using sharded simulation - a set of smaller instructions that collectively convey the same information as the original instruction. In this setting, models tend to over-rely on incorrect assumptions they made in earlier turns.

Multiple methods have been proposed to overcome the limitations of long contexts and unlock further scaling of test-time compute, including context compaction, summarization, and external memory modules \citep{anthropic2025context_engineering, liu2025deepseekv3_2, yan2025inftythink, memagent}. A parallel line of work takes advantage of problems that can be split into isolated subproblems by delegating them to recursive self-calls, which may improve efficiency and naturally allow for maintaining compact context representations \citep{yang2025pencil, jin2025learningpromisescalinglanguage, ning2024skeletonofthought, zheng2025parallelr1}.

\vspace{-10px}
\section{Experiments}\label{sect:experiments}
\vspace{-10px}

\subsection{Setup}\label{sect:exp_setup}
\vspace{-5px}

This section presents experiments designed to answer a key question: Can a model solve an isolated subproblem as effectively when surrounded by irrelevant context as it does in isolation? This inquiry is motivated by two observations. First, complex reasoning tasks can often be decomposed into independent subtasks that do not require global context. Second, long-running agents increasingly operate within broad, general contexts that inevitably contain details irrelevant to specific user queries or subtasks.

To evaluate the model's ability to reason about the same problems under different context conditions, we compare the following setups:
\begin{itemize}
    \item \textbf{Baseline}: a single user message containing a problem with a standard prompt.
    \item \textbf{Subtask}: a single user message containing two independent problems to solve.
    \item \textbf{Long Input}: a single user message containing a long chunk of irrelevant data (Shakespeare's plays from the \cite{char-rnn}) followed by a problem with a standard prompt.
    \item \textbf{Multi-turn}: a multi-turn chat history, where each user message asks for solving a new problem with the Baseline prompt. We only evaluate the second turn.
   
\end{itemize}

These synthetic context conditions are controlled experimental probes designed to isolate and characterize the effect of redundant context. Please refer to Appendix \ref{app:setup_design} for the discussion on the setup design.

\begin{table}[h]
\begin{tabular}{lcccccccc}
\toprule
\multirow{2}{*}{\textbf{Model}} & \multicolumn{2}{c}{\textbf{Baseline}} & \multicolumn{2}{c}{\textbf{Subtask}} & \multicolumn{2}{c}{\textbf{Long Input}} & \multicolumn{2}{c}{\textbf{Multi-turn}} \\
\cmidrule(lr){2-3} \cmidrule(lr){4-5} \cmidrule(lr){6-7} \cmidrule(lr){8-9}
 & \textbf{Acc.} & \textbf{Tok.} & \textbf{Acc.} & \textbf{Tok.} & \textbf{Acc.} & \textbf{Tok.} & \textbf{Acc.} & \textbf{Tok.} \\
\midrule
\multicolumn{9}{c}{\textbf{IMOAnswerBench}} \\
\midrule
Qwen3.5-27B & 74.5 & 28,771 & \cellcolor{red!16}62.4 & \cellcolor{red!30}20,165 & \cellcolor{red!9}67.8 & \cellcolor{red!43}16,415 & \cellcolor{red!10}67.0 & \cellcolor{red!40}17,404 \\
Gemma 4 31B & 72.7 & 9,240 & \cellcolor{red!40}52.8 & \cellcolor{red!30}5,635 & \cellcolor{red!9}67.0 & \cellcolor{red!43}6,252 & \cellcolor{red!10}71.0 & \cellcolor{red!40}7,530 \\
gpt-oss-120b & 73.8 & 24,180 & \cellcolor{red!13}64.0 & \cellcolor{red!28}17,408 & \cellcolor{red!13}64.0 & \cellcolor{red!51}11,876 & \cellcolor{red!6}69.3 & \cellcolor{red!18}19,831 \\
Gemini 3 Flash Preview & 82.8 & 23,090 & \cellcolor{red!19}67.0 & \cellcolor{red!41}13,653 & \cellcolor{red!5}80.3 & \cellcolor{red!14}19,879 & \cellcolor{red!5}82.5 & \cellcolor{red!6}21,693 \\
Kimi K2 Thinking & 74.8 & 29,615 & \cellcolor{red!13}65.0 & \cellcolor{red!34}19,630 & \cellcolor{red!5}70.8 & \cellcolor{red!21}23,380 & \cellcolor{red!5}72.8 & 30,421 \\
\midrule
\multicolumn{9}{c}{\textbf{GPQA-Diamond}} \\
\midrule

Qwen3.5-27B & 85.1 & 12,364 & 83.6 & \cellcolor{red!29}7,279 & 82.1 & \cellcolor{red!31}6,939 & 83.8 & \cellcolor{red!43}4,722 \\
Gemma 4 31B & 84.1 & 3,638 & \cellcolor{red!6}79.8 & \cellcolor{red!27}2,219 & 82.6 & \cellcolor{red!31}2,047 & \cellcolor{red!8}78.5 & \cellcolor{red!44}1,353 \\
gpt-oss-120b & 78.3 & 10,762 & \cellcolor{red!21}64.0 & \cellcolor{red!21}7,466 & \cellcolor{red!31}57.6 & \cellcolor{red!46}3,711 & \cellcolor{red!9}72.5 &
\cellcolor{red!41}4,476 \\
Gemini 3 Flash Preview & 89.6 & 12,927 & 89.4 & \cellcolor{red!27}7,918 & 89.6 & \cellcolor{red!11}10,922 & 88.7 & \cellcolor{red!52}3,334 \\
Kimi K2 Thinking & 81.6 & 11,121 & \cellcolor{red!16}70.7 & \cellcolor{red!22}7,635 & \cellcolor{red!5}78.0 & \cellcolor{red!22}7,619 & \cellcolor{red!13}73.2 &
\cellcolor{red!41}4,669 \\
\bottomrule
\end{tabular}
\centering
\caption{Model performance on IMOAnswerBench and GPQA-Diamond. Accuracy and average number of generated reasoning tokens are shown. Background color represents the \textcolor{red!80}{relative change} from the baseline values.}
\label{tab:main_exp}
\end{table}

For our main experiments, we evaluate the following models on IMOAnswerBench \citep{imobench} and GPQA-Diamond~\citep{rein2023gpqagraduatelevelgoogleproofqa}: Qwen3.5-27B \citep{qwen35blog}, gpt-oss-120b \citep{gpt-oss}, Gemma 4 31B \citep{gemma4}, Gemini 3 Flash Preview \citep{gemini3flash}, and Kimi K2 Thinking \citep{kimik2thinking}. For each model, we report both accuracy and the number of reasoning tokens generated. IMOAnswerBench consists of 400 Olympiad-level math problems with short verifiable answers (including mathematical expressions): we use Gemini 3.1 Pro Preview \citep{gemini31} with the original prompt for AnswerAutoGrader\citep{imobench} as a judge for automatic evaluation. For GPQA-Diamond, we report results averaged across 2 runs.

\begin{figure}[t]
    \centering
    \includegraphics[width=0.46\linewidth]{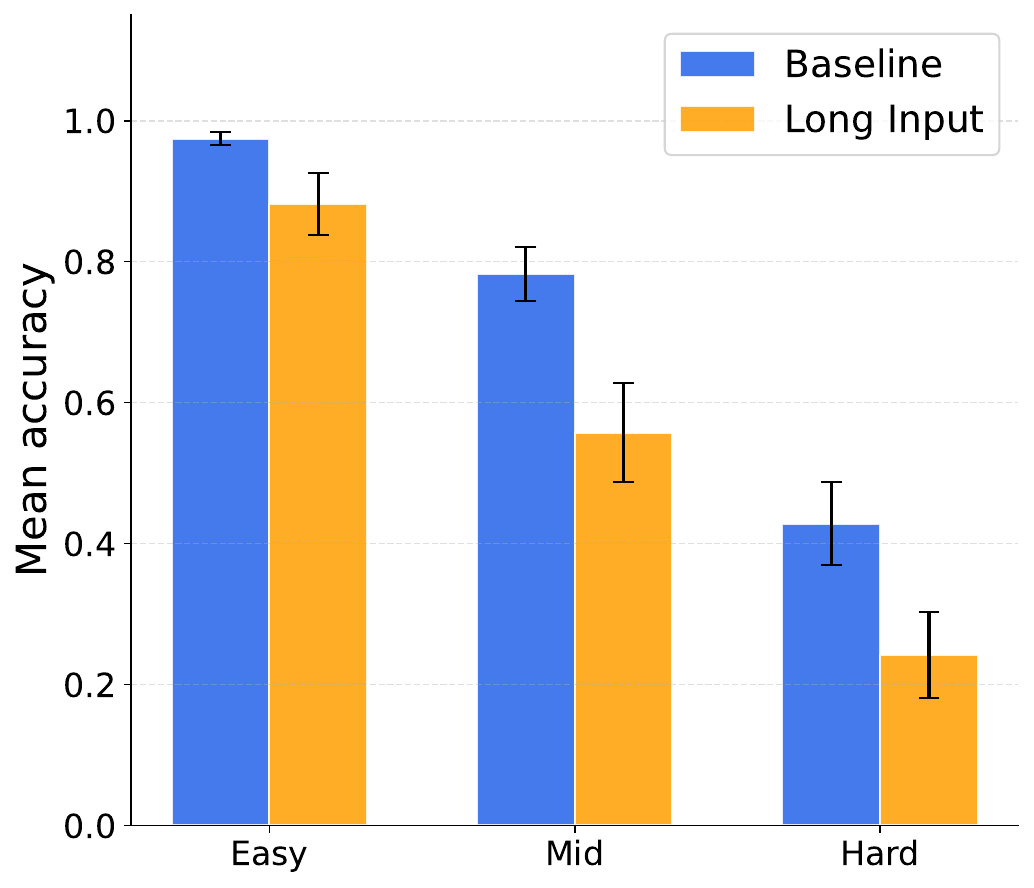}
    \includegraphics[width=0.46\linewidth]{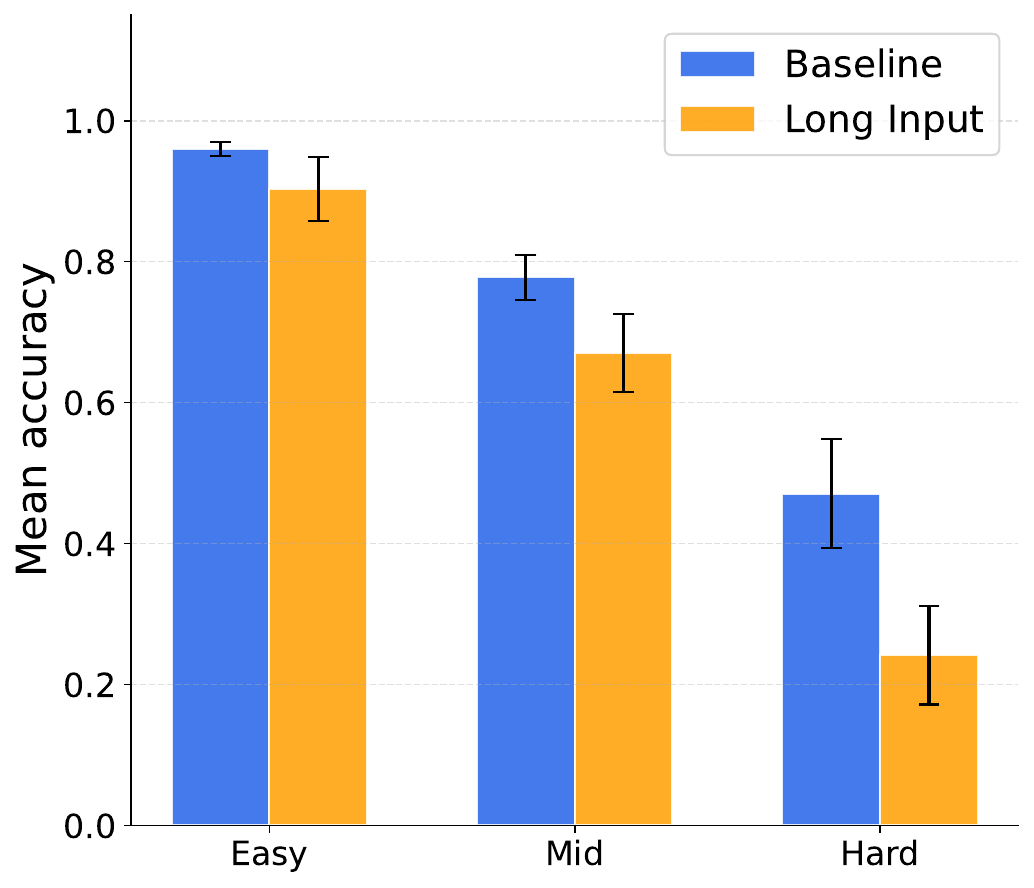}

    \caption{Accuracy comparison between Baseline and Long Input conditions across task difficulty levels (Easy, Mid, Hard) on a subset of IMOAnswerBench. \textbf{Left:} Gemma 4 31B, \textbf{right:} gpt-oss-120b.}
    \label{fig:difficulty_hist}
    \vspace{-10px}
\end{figure}

\begin{figure}[!b]
    \centering
    \vspace{-5px}
    \includegraphics[width=0.46\linewidth]{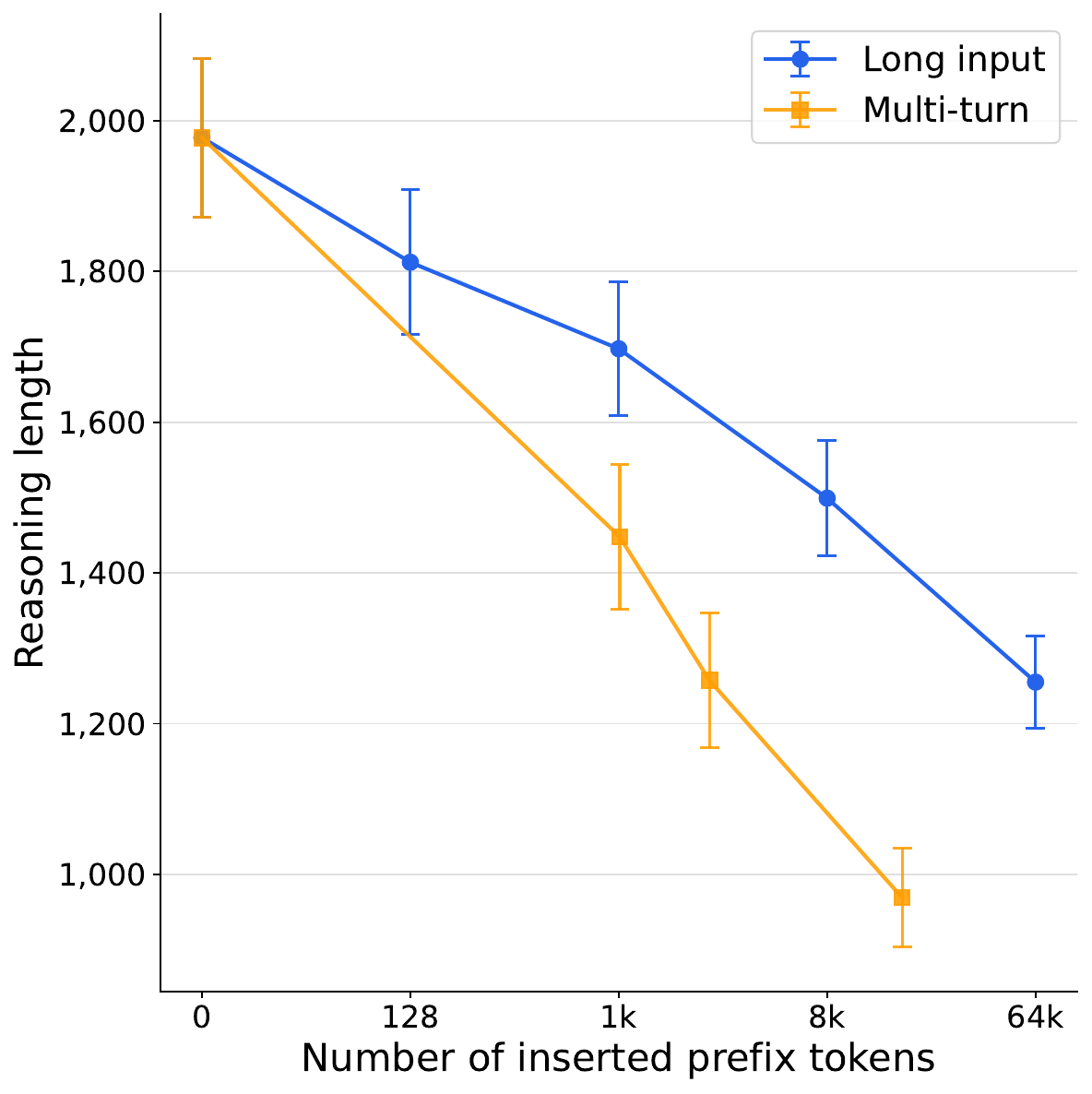}
    \includegraphics[width=0.46\linewidth]{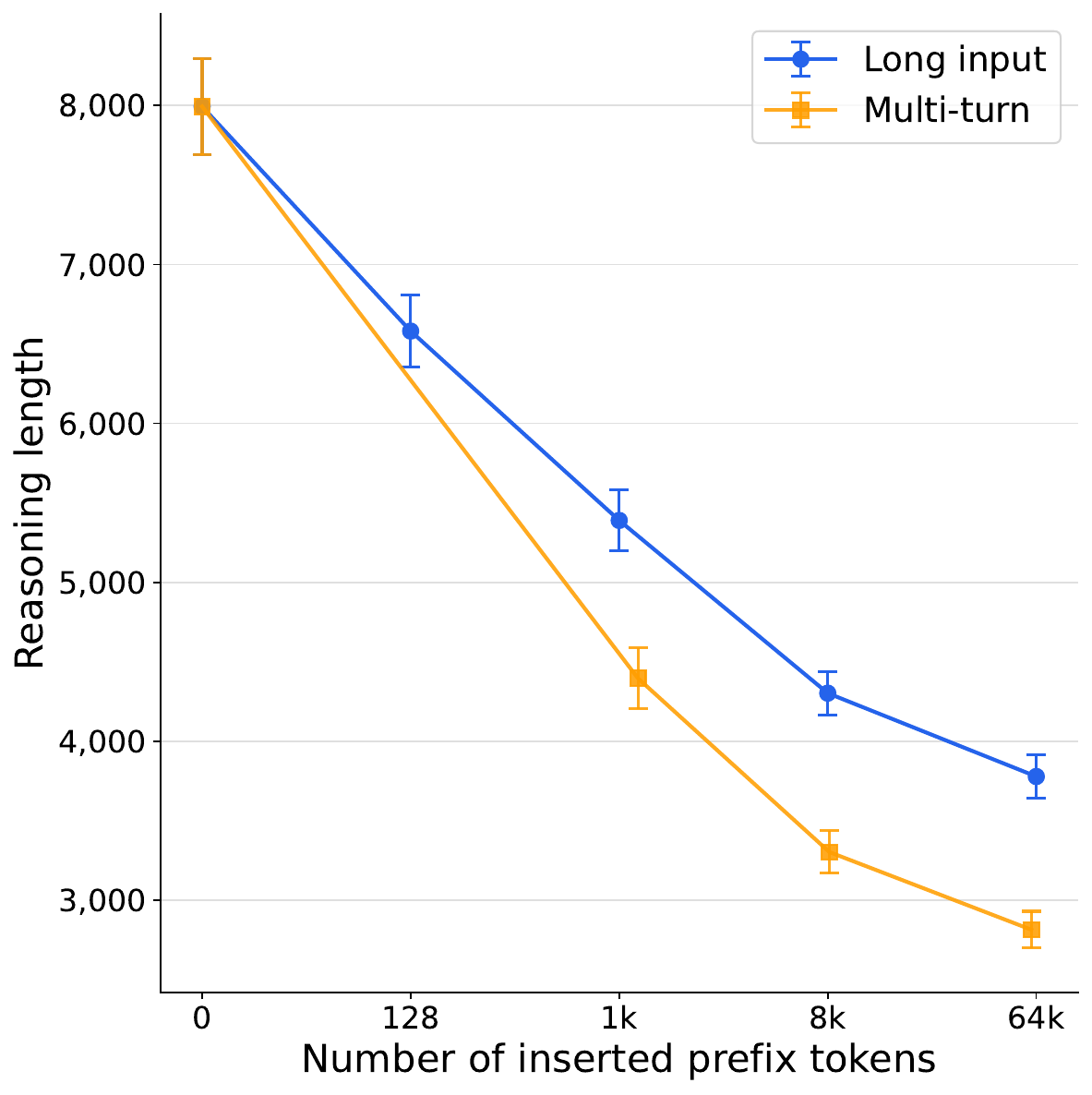}

    \caption{Average reasoning length on MATH-500 under varying number of inserted tokens in Long Input and Multi-turn setups. \textbf{Left:} Gemma 4 31B, \textbf{right:} Qwen3.5-27B.}
    \label{fig:vary_inserted_tokens}
    \vspace{-10px}
\end{figure}

For Subtask scenario, model solves two independent problems within the same query. We report average accuracy of solving both problems and the number of reasoning tokens divided by two, representing the average number of reasoning tokens spent on a single task. Please refer to Appendix~\ref{app:exp_details} for additional details on the experimental setup and to Appendix~\ref{app:lcb_code_experiment} for additional evaluation on code generation tasks.

\vspace{-5px}
\subsection{Results}\label{sect:results}
\vspace{-5px}

Results are presented in Table \ref{tab:main_exp}. Interestingly, we observe a slight performance drop in all Subtask and Long Input scenarios for IMOAnswerBench: degradation of 12\% for Qwen3.5-27B, 9\% for gpt-oss-120b, 15\% for Gemini 3 Flash Preview and 9\% for Kimi K2 Thinking. We report a detailed evaluation of the Subtask scenario in Appendix \ref{app:subtask_evaluation_details}.

Importantly, all models covered produce much shorter reasoning traces under different non-baseline context conditions for both benchmarks, generating up to 74\% fewer reasoning tokens on average for the same problems ($p < 10^{-10}$ with paired Wilcoxon signed-rank test, for all models under the Long Input condition on IMOAnswerBench). Please see Figure \ref{fig:main_exp_task_wise} for task-wise comparison of the number of generated tokens.

Overall, we observe a significant distributional shift that causes symptoms such as shorter reasoning and degraded performance.

\vspace{-5px}
\subsection{Additional experiments}\label{sect:exps_additional}
\vspace{-5px}

To further examine how task difficulty interacts with the observed performance degradation, we sampled 50 runs per task for 30 selected problems from IMOAnswerBench under both Baseline and Long Input conditions, grouping tasks into easy, medium, and hard buckets based on Baseline accuracy. We find that the accuracy drop under Long Input is disproportionately concentrated on medium and hard problems, see Figure \ref{fig:difficulty_hist}. This suggests that the reasoning shift induced by additional context is relatively benign for straightforward problems but becomes increasingly costly as task difficulty rises.

Importantly, as shown in Figure \ref{fig:vary_inserted_tokens}, increasing the number of input context leads to a gradual and consistent reduction in reasoning length in both Long Input and Multi-turn scenarios. For example, for Qwen3.5-27B, even short distractions (hundreds of tokens) may be enough to reduce the average reasoning length by 18\%, while further increasing the prompt size reduces reasoning by 50\%. In the Multi-turn setup, adding more turns to the chat history also gradually reduces reasoning length. 

Further, we additionally test whether the domain relevance of inserted prefix under Long Input scenario would affect the observed phenomenon. Results are presented in Appendix \ref{app:long_input_relevant}.

\begin{figure}[t]
    \centering
    \begin{subfigure}[b]{0.45\textwidth}
        \centering
        \includegraphics[width=\textwidth]{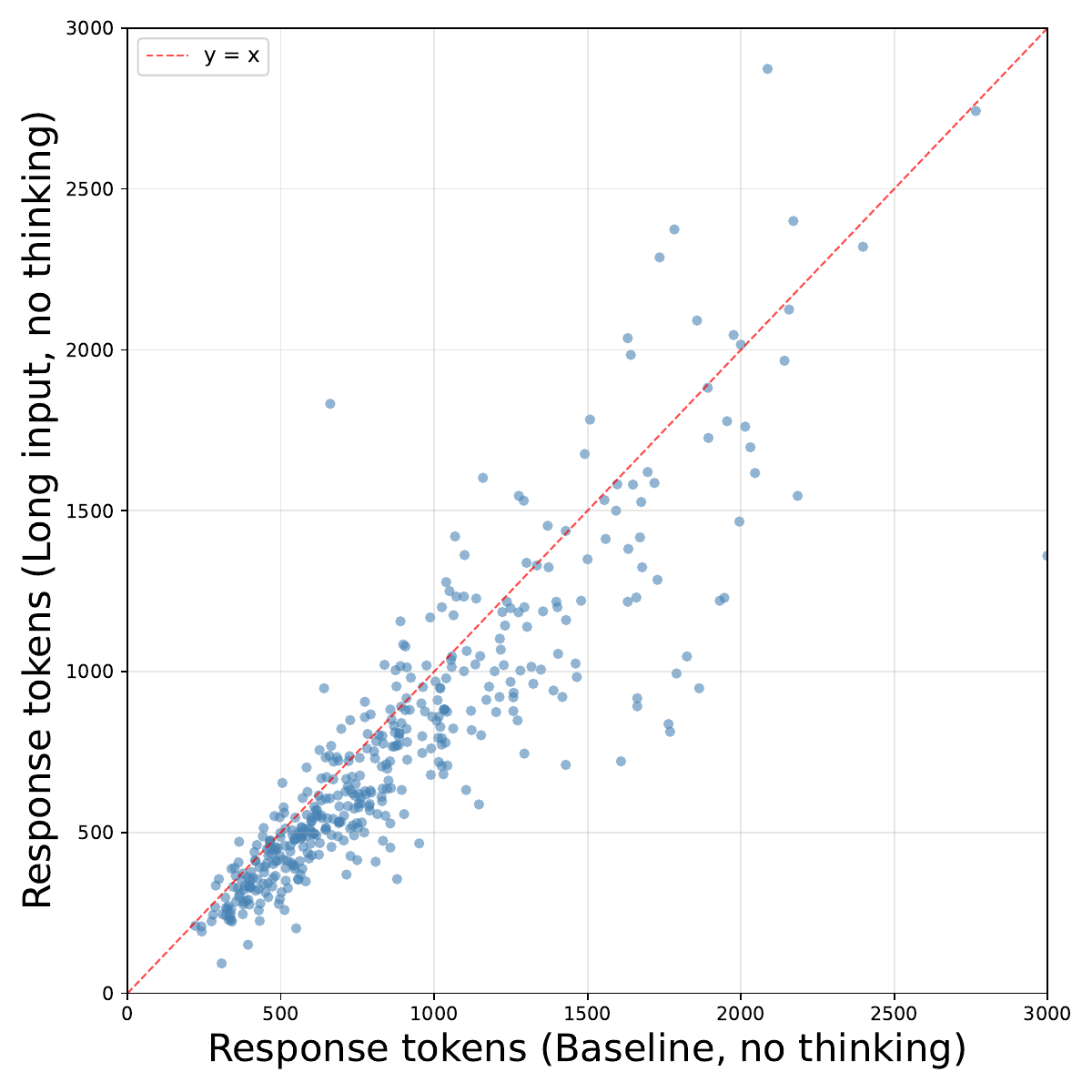}
        \caption{Number of response tokens, \textbf{no thinking} mode.}
        \label{fig:plot1}
    \end{subfigure}
    \hfill
    \begin{subfigure}[b]{0.45\textwidth}
        \centering
        \includegraphics[width=\textwidth]{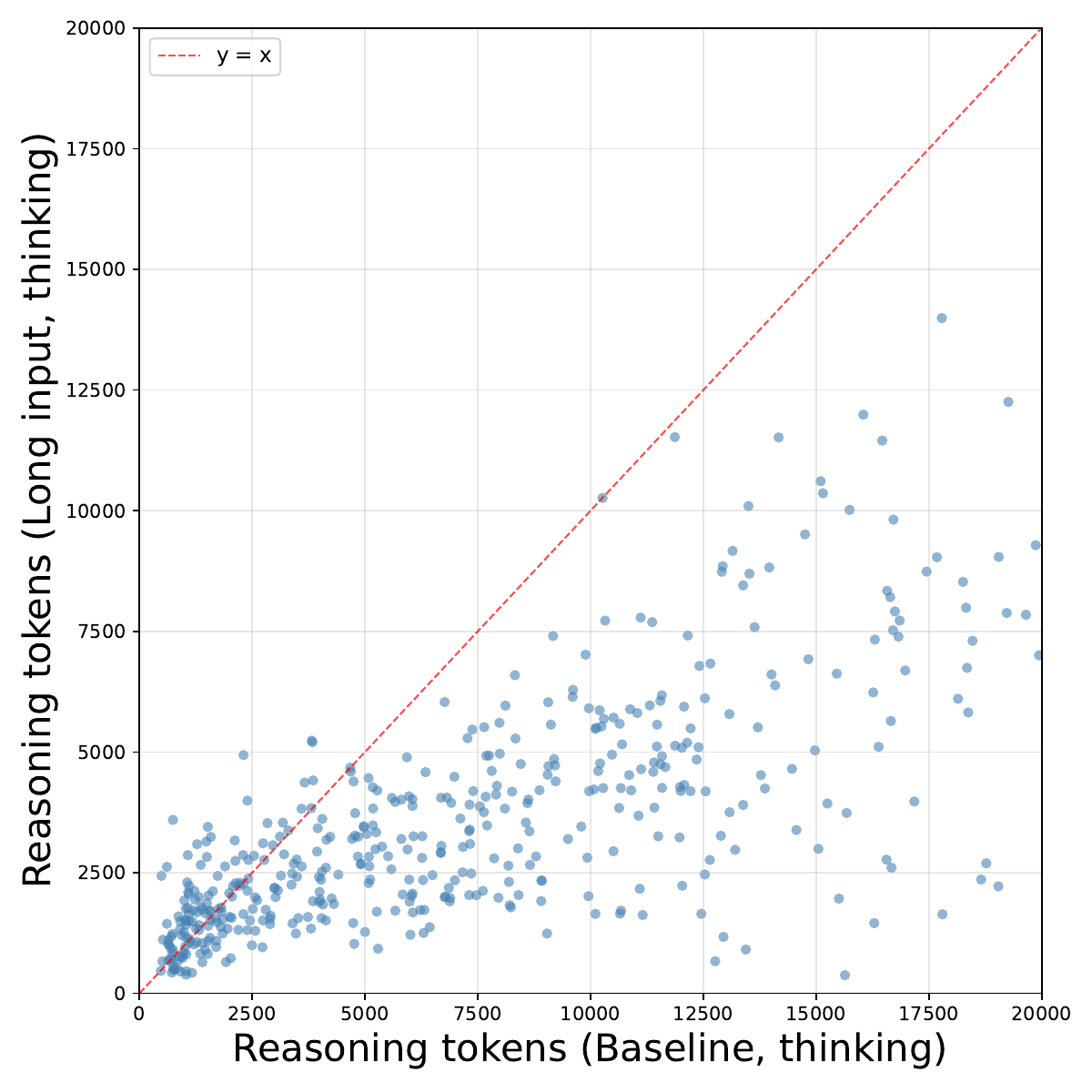}
        \caption{Number of reasoning tokens, \textbf{thinking} mode.}
        \label{fig:plot2}
    \end{subfigure}
    \caption{Number of generated tokens for Qwen3.5-27B for each MATH-500 task. X-axis: Baseline, Y-Axis: Long Input.}
    \label{fig:no_think_vs_think}
\end{figure}

\begin{table}[b]
\centering

\begin{tabular}{lcccccccc}
\toprule
\multirow{2}{*}{\textbf{Model}} & \multicolumn{2}{c}{\textbf{Baseline}} & \multicolumn{2}{c}{\textbf{Subtask}} & \multicolumn{2}{c}{\textbf{Long Input}} & \multicolumn{2}{c}{\textbf{Multi-turn}} \\
\cmidrule(lr){2-3} \cmidrule(lr){4-5} \cmidrule(lr){6-7} \cmidrule(lr){8-9}
 & \textbf{Acc.} & \textbf{Tok.} & \textbf{Acc.} & \textbf{Tok.} & \textbf{Acc.} & \textbf{Tok.} & \textbf{Acc.} & \textbf{Tok.} \\
\midrule
Olmo-3-7B-Instruct & 91.2 & 925 & \cellcolor{red!5}89.1 & 1,139 & \cellcolor{red!5}89.8 & 1,162 & \cellcolor{red!5}90.5 & 1,089 \\
Olmo-3-7B-Think-SFT & 94.5 & 2,671 & \cellcolor{red!5}90.2 & \cellcolor{red!32}1,829 & \cellcolor{red!5}91.4 & \cellcolor{red!29}1,908 & \cellcolor{red!5}89.1 & \cellcolor{red!40}1,611 \\
Olmo-3-7B-Think-DPO & 93.5 & 3,051 & \cellcolor{red!5}89.7 & \cellcolor{red!25}2,286 & \cellcolor{red!5}91.0 & \cellcolor{red!20}2,437 & \cellcolor{red!5}88.7 & \cellcolor{red!22}2,395 \\
Olmo-3-7B-Think & 95.2 & 3,664 & \cellcolor{red!5}92.1 & \cellcolor{red!22}2,851 & \cellcolor{red!5}93.2 & \cellcolor{red!22}2,859 & \cellcolor{red!5}92.2 & \cellcolor{red!24}2,788 \\
\bottomrule
\end{tabular}
\caption{Model performance with accuracy and average number of generated reasoning tokens, MATH-500. For Instruct model, the average number of response tokens is reported. Background color represents the \textcolor{red!80}{relative change} from the baseline values.}
\label{tab:post_training}
\end{table}

\vspace{-5px}
\subsection{Relation to thinking mode}\label{sect:exps_thinking_mode}
\vspace{-5px}

We also conduct additional experiments to determine whether this effect is specific to reasoning mode. For this purpose, we evaluate Qwen3.5-27B, a model with integrated thinking and non-thinking modes, on the MATH-500 benchmark \citep{lightman2023lets_verify_step_by_step} using the Long Input scenario (see Figure \ref{fig:no_think_vs_think} for a task-wise comparison of the number of generated tokens). For non-thinking mode, we report the response length. Qwen3.5-27B demonstrates a 19\% reduction in response length in non-thinking mode (1664 tokens for Baseline vs. 1348 tokens for Long Input) and a 53\% reduction in reasoning length (8003 tokens for Baseline vs. 3762 tokens for Long Input). While both thinking and non-thinking modes exhibit statistical changes under different context conditions, we observe that the described phenomenon is markedly more pronounced in thinking mode.

Focusing on reasoning mode, we analyze checkpoints from different stages of post-training to examine how they affect reasoning shifts under varying context conditions. For this purpose, we evaluate different checkpoints of the Olmo3 model \citep{olmo3}. In short, we observe the described phenomenon across all reasoning checkpoints. The results are presented in Table \ref{tab:post_training}.

\vspace{-10px}
\section{Analysis}\label{sect:analysis}
\vspace{-10px}

Our experiments in Section \ref{sect:experiments} demonstrate a significant difference in the number of reasoning tokens generated for the same problems under different context conditions. In this section, we analyze the differences between these traces. Please refer to Appendix \ref{app:exp_details_analysis} for the details on experimental setup.

We inspect reasoning traces produced by Qwen3-32B, Qwen3.5-27B and Gemma 4 31B for MATH-500 problems. For clarity of presentation, we report here the results for Qwen3-32B, with references to the additional experiments with other models.

First, we manually inspect the traces to find any evidence of how the context might affect the reasoning - we did not find any indication that the model became confused by the query or failed to understand the task. Specifically, when investigating the reasoning traces produced by Qwen3-32B under Long Input prompts, we find only brief references to the irrelevant prompt part; we report several examples in Appendix \ref{app:long_input_refs}. These are all brief, dismissive acknowledgments: the model notices the prefix, labels it irrelevant, and moves on immediately. While additional context clearly affects the model’s internal computations, we found no evidence that the inserted content influenced the model at a semantic level beyond simple acknowledgment.

\vspace{-5px}
\subsection{Does the model arrive at the answer faster?}\label{sect:analysis_answer_position}
\vspace{-5px}

We compare whether Baseline traces need more time to arrive at the answer or are longer because they exhibit stronger verification patterns after a candidate answer is found. For this purpose, we identify the position within each trace where the first candidate answer is stated. Results are reported in Table \ref{tab:first_answer_position}. 
Interestingly, while some differences are observed (-12.4\% for Long Input and -5.5\% for Multi-turn compared to Baseline), those differences are significantly lower than the average reduction of all reasoning traces, meaning that the parts of reasoning traces from the moment when first candidate answer is found until the end of the reasoning trace were reduced much more intensively (-36.1\% for Long Input and -21.1\% for Multi-turn compared to Baseline). Other models share the same trend.

\vspace{-5px}
\subsection{What is different in the reasoning dynamics?}\label{sect:analysis_thought_anchors}
\vspace{-5px}

To further investigate the nature of the significant differences in reasoning length, we adopt the framework from \citep{venhoff2025understanding} to perform sentence-level analysis of long reasoning traces. We classify each sentence within the traces into the following categories: problem setup, plan generation, fact retrieval, active computation, result consolidation, uncertainty management, and final answer emission. We investigate the transition matrices of these labels within the reasoning traces. Figure \ref{fig:heatmap} demonstrates the difference in transition heatmaps for the Baseline and Long Input setups. We note the absolute largest value in this heatmap: the transition from final answer emission to the end of the thinking trace (57\% for Baseline vs. 68\% for Long Input), which may indicate a significant behavioral difference: once the final answer is stated,  Long Input traces finish more often, whereas Baseline traces have a greater probability of initiating additional self-checks.

\vspace{-5px}
\subsection{Does additional context suppress self-verification?}\label{sect:analysis_resampling}
\vspace{-5px}

\begin{wraptable}{r}{0.48\textwidth} 
\centering
\begin{tabular}{lccc}
\toprule
\textbf{Token} & \textbf{0} (\textbf{Baseline}) & \textbf{128} & \textbf{16k} \\
\midrule
</think> & 21\% & 26\% & 46\% \\
Wait & 11\% & 10\% & 5\% \\
Alternatively & 17\% & 11\% & 5\% \\
But & 46\% & 38\% & 20\% \\
Maybe & 23\% & 17\% & 9\% \\
\bottomrule
\end{tabular}
\caption{Resampling from the same reasoning traces but under varying numbers of inserted prompt tokens in the Long Input setup. The table presents the ratio of traces containing the end of reasoning or self-verification tokens.}
\label{tab:resampling}
\end{wraptable}

The changes in self-checks frequency may be influenced by the content of the reasoning traces themselves rather than by the context conditions. To test this, we conduct a resampling experiment: we take the traces produced under Long Input and Multi-turn conditions, remove the final 64 tokens from the reasoning part, and resample the same number of tokens under different context conditions: Baseline, Long Input (128) (a condition similar to Long Input but with only 128 inserted tokens),  Long Input (16k) and Multi-turn.  We resample each trace with the same sampling parameters as the original traces - recommended for reasoning tasks. For the resampled traces, we compare the ratio of finished traces. The results are presented in Table \ref{tab:resampling}.

We observe an interesting phenomenon: when resampling from almost finished reasoning samples under different context conditions, all non-baseline context conditions have higher rates of finished reasoning traces and lower frequencies of words using during self-verification and uncertainty management. The ratio of finished traces for Long Input is 46\%, compared to 21\% for Baseline, which is also accompanied by reduced frequencies of words used during self-verification and uncertainty management, such as "wait," "alternatively," and "but." Please refer to Appendix \ref{app:resampling_samples} for the examples of the resampled samples. In Table \ref{tab:resampling_qwen_gemma} we additionally report the results of a similar resampling experiment for other models and Multi-turn setup, which demonstrates the same trend.

\vspace{-5px}
\subsection{Does additional context affect verbalized self-confidence?}\label{sect:analysis_verbalised_confidence}
\vspace{-5px}

Recent studies demonstrate that language models use confidence to drive their behavior \citep{kumaran2026causal, kumaran2026detect_own_errors}, which raises a question: do different context conditions affect models' confidence in their reasoning? To test this we ask the model to verbalize the confidence in their partial solutions under different context conditions (similar to our resampling experiment). For confidence evaluation, we remove last 64 tokens of the reasoning trace and perform an intervention with a self-confidence prompt from \cite{kumaran2026how_llm_compute_confidence} and use greedy decoding for generating the verbalized confidence score. Results are presented in Table \ref{tab:self_confidence}. We observe that the same reasoning traces produced from both the Baseline and non-baseline setups receive lower confidence scores under the Baseline setup on average, which is consistent with the changes in rates of self-verification in our resampling experiment. In Appendix \ref{app:exp_details_analysis}, we report additional results on verbalized confidence, evaluations with different prompts and reasoning prefixes, all leading to similar conclusions. These results suggest that the ability of the model to estimate its confidence might not be robust to OOD context conditions.

\begin{table}[b]
\centering

\begin{tabular}{lcccc}
\toprule
\textbf{Trace} & \textbf{Baseline} & \textbf{Long Input(128)} & \textbf{Long Input(16k)} & \textbf{Multi-turn} \\
\midrule
\textbf{Baseline}       & 62.0\%     & 68.0\%          & 90.4\%           & 78.4\% \\
\textbf{Long Input}   & 41.2\%     & 48.2\%          & 85.4\%           & 64.4\% \\
\textbf{Multi-turn}   & 46.8\%     & 55.5\%           & 86.7\%           & 70.5\% \\
\midrule
\end{tabular}
\caption{The same reasoning traces receive higher self-confidence scores under non-baseline context conditions. Table represents the ratio of traces with the highest self-confidence scores. Rows represent how traces were generated, columns represent the context used during self-confidence evaluation. Qwen3-32B, MATH-500.}
\label{tab:self_confidence}
\end{table}

\vspace{-5px}
\subsection{Would the model benefit from longer thinking under context pressure?}\label{sect:analysis_wait_internvetion}
\vspace{-5px}

Our results suggest that, for the same reasoning prefixes, different context conditions may suppress high-level reasoning patterns, such as self-verification and uncertainty management, which make reasoning traces significantly shorter. That raises a question: would the model be able to recover the Baseline-level accuracy if forced to think longer?

We took reasoning traces generated under the Long Input condition for IMOAnswerBench problems and, at the end of that reasoning, intervened with “Wait, let's check everything again.” Similar to our resampling experiment, we considered interventions under different context conditions: Baseline and Long Input. Note that the interventions were applied to the same reasoning traces, which were produced under the Long Input setup. Results are presented in Table \ref{tab:wait_intervention}.

\begin{table}[t]
    \centering

    \small
    \setlength{\tabcolsep}{3pt}
    \renewcommand{\arraystretch}{1.1}

    \begin{tabular}{@{}l *{8}{c}@{}}
        \toprule
        & \multicolumn{2}{c}{Long Input}
        & \multicolumn{2}{c}{\makecell{``Wait'' intervention\\(Long Input context)}}
        & \multicolumn{2}{c}{\makecell{``Wait'' intervention\\(Baseline context)}}
        & \multicolumn{2}{c}{Baseline} \\
        \cmidrule(lr){2-3} \cmidrule(lr){4-5}
        \cmidrule(lr){6-7} \cmidrule(l){8-9}
        Model & Acc. & Tokens & Acc. & Tokens
              & Acc. & Tokens & Acc. & Tokens \\
        \midrule
        Gemma 4 31B
            & \cellcolor{red!8}67.0 & \cellcolor{red!32}6,252
            & \cellcolor{red!6}68.3 & \cellcolor{red!13}8,058
            & \cellcolor{red!1}71.8 & 10,301
            & 72.7 & 9,240 \\
        Qwen3.5-27B
            & \cellcolor{red!9}67.8 & \cellcolor{red!43}16,415
            & \cellcolor{red!8}68.5 & \cellcolor{red!37}18,191
            & \cellcolor{red!7}69.3 & \cellcolor{red!31}19,865
            & 74.5 & 28,771 \\
        \bottomrule
    \end{tabular}
    \caption{Effect of a “Wait” intervention on IMOAnswerBench accuracy and average reasoning-token count. The same reasoning traces generated under Long Input are continued after the intervention with the irrelevant context either retained or removed.}
\label{tab:wait_intervention}
\end{table}

We can observe that, for both models considered, the "wait" intervention reduces the gap between the Long Input and Baseline conditions in terms of both accuracy and the number of reasoning tokens. Importantly, both models benefit from additional reasoning under both context conditions. However, for both models, the intervention under the Baseline context is more effective for the same reasoning traces: the models tend to review their solutions for longer and produce better final answers.

Overall, we can observe that additional context affects self-confidence reported by the model, suppresses the proportion of traces in which additional self-checks are initiated and limits the effectiveness of these self-verifications.

\vspace{-10px}
\section{Mitigation approaches}\label{sect:mitigation}
\vspace{-10px}

In this section, we study potential approaches for mitigating the observed phenomenon as an additional diagnostic analysis.
We discuss two methods: (1) prompting models to engage in more deliberate reasoning, and (2) fine-tuning models on a mixture of data that includes non-baseline context conditions. 

\vspace{-5px}
\subsection{Prompting}\label{sect:prompting}
\vspace{-5px}
We first investigate whether the observed phenomenon can be mitigated through prompting alone. To test this, we evaluate multiple models under the same conditions as in the main experiment but with a prompt explicitly requesting more deliberate and thorough reasoning. Results are presented in Table~\ref{tab:max_effort}. We experimented with several prompt formulations and report results for the best-performing one - the maximum reasoning effort prompt used for DeepSeek v4 \citep{deepseekv42026}, see Appendix \ref{app:exp_details}.

Overall, reasoning models show limited sensitivity to prompts requesting additional reasoning effort. In the baseline setting, the max effort prompt produces modestly longer reasoning traces (10–15\% more tokens on average) with comparable or slightly improved accuracy. However, under non-baseline context conditions, the same prompt yields a similar relative increase in trace length while preserving the same shortening rate observed without the intervention. For example, gpt-oss-120b generates approximately 14\% more tokens with the max effort prompt under the Long Input condition (13{,}494 vs.\ 11{,}876), yet this still represents a 51\% reduction from the max effort baseline (27{,}525 tokens), nearly identical to the 51\% reduction observed with the standard prompt (24{,}180 to 11{,}876). These results suggest that prompting alone is insufficient to counteract the reasoning shift induced by non-baseline context conditions.

\begin{table}[b]
\centering

\begin{tabular}{ll cc cc}
\toprule
& & \multicolumn{2}{c}{\textbf{Baseline}}
  & \multicolumn{2}{c}{\textbf{Long Input}} \\
\cmidrule(lr){3-4} \cmidrule(lr){5-6}
\textbf{Model} & \textbf{Prompt}
& \textbf{Acc.} & \textbf{Tokens}
& \textbf{Acc.} & \textbf{Tokens} \\
\midrule
\multirow{2}{*}{Qwen3.5-27B}
& Standard   & 74.5 & 28,771
& \cellcolor{red!9}67.8 & \cellcolor{red!43}16,415 \\
& Max effort & 74.4 & 28,755
& \cellcolor{red!5}70.5 & \cellcolor{red!44}15,960 \\
\midrule
\multirow{2}{*}{Gemma 4 31B}
& Standard   & 72.7 & 9,240
& \cellcolor{red!8}67.0 & \cellcolor{red!32}6,252 \\
& Max effort & 73.8 & 10,624
& \cellcolor{red!4}70.9 & \cellcolor{red!32}7,198 \\
\midrule
\multirow{2}{*}{gpt-oss-120b}
& Standard   & 73.8 & 24,180
& \cellcolor{red!13}64.0 & \cellcolor{red!51}11,876 \\
& Max effort & 73.9 & 27,525
& \cellcolor{red!16}61.8 & \cellcolor{red!51}13,494 \\
\bottomrule
\end{tabular}
\caption{Effect of max reasoning effort prompt on IMOAnswerBench.}
\label{tab:max_effort}
\end{table}

\vspace{-5px}
\subsection{Fine-tuning}\label{sect:sft}
\vspace{-5px}
We also explore a simple fine-tuning approach aimed at preventing models from reducing reasoning length when irrelevant information is present in the context. Our analysis in Section \ref{sect:analysis} indicates that the mechanisms responsible for self-verification are not robust to out-of-distribution inputs, although other explanations remain possible (e.g., general architectural limitations in processing long contexts). The fine-tuning experiment is intended to test whether the observed behavioral shift has a substantial distributional component.

Results in Table~\ref{tab:post_training} suggest that the phenomenon may emerge during the early stages of reasoning-specific post-training. Motivated by this finding, we focus on mitigating the effect at the SFT stage. A straightforward approach would be to augment all SFT training samples with irrelevant context. Instead, we try to adapt an already SFT-trained reasoning model using a relatively small amount of additional data.

\textbf{Training data.} 
Starting from the \texttt{allenai/Dolci-Think-SFT-7B} dataset \citep{olmo3}, we augment training samples with additional irrelevant context. We use only 10\% of the dataset for our experiment. For each sample, one of several augmentation strategies is selected uniformly at random: (1) keeping the original sample unchanged, (2) prepending a long irrelevant Shakespeare text fragment, (3) prepending a single irrelevant user-assistant interaction, or (4) prepending multiple irrelevant user-assistant turns. We limit the number of inserted irrelevant context to 16k tokens. For multi-turn augmentation, we use user queries from the \texttt{allenai/Dolci-Think-DPO-7B} dataset and generate assistant responses using Olmo-3-7B-Think-SFT, producing synthetic prompt-response pairs. Please refer to Appendix \ref{app:fine_tuning} for additional experiment details.

\begin{table}[t]
\centering

\small
\begin{tabular}{@{}lcccccccc}
\toprule
\multirow{2}{*}{\textbf{Model}} 
& \multicolumn{2}{c}{\textbf{Baseline}} 
& \multicolumn{2}{c}{\textbf{Multi-turn(16)}} 
& \multicolumn{2}{c}{\textbf{Multi-turn(32)}} 
& \multicolumn{2}{c}{\textbf{Long Input(32)}} \\
\cmidrule(lr){2-3}
\cmidrule(lr){4-5}
\cmidrule(lr){6-7}
\cmidrule(lr){8-9}
& \textbf{Acc.} 
& \textbf{Tok.} 
& \textbf{Acc.} 
& \textbf{Tok.} 
& \textbf{Acc.} 
& \textbf{Tok.} 
& \textbf{Acc.} 
& \textbf{Tok.} \\
\midrule

Olmo-3-7B-Think-SFT 
& 94.5
& 2,671
& \cellcolor{red!5}92.2
& \cellcolor{red!33}1,781
& \cellcolor{red!12}89.1
& \cellcolor{red!40}1,611
& \cellcolor{red!7}91.4
& \cellcolor{red!29}1,908 \\

Olmo-3-7B-Think-SFT-Ours
& 96.1
& 2,869
& \cellcolor{red!2}94.5
& \cellcolor{red!3}2,674
& \cellcolor{red!6}89.8
& \cellcolor{red!4}2,606
& \cellcolor{red!2}93.8
& \cellcolor{red!1}2,806 \\

\bottomrule
\end{tabular}
\caption{Model performance with accuracy and average number of generated reasoning tokens under different context conditions on MATH-500.}
\label{tab:multiturn_sft_math500}
\end{table}

\textbf{Evaluation.} We evaluate the resulting model on MATH-500 under the Multi-turn(16), Multi-turn(32), and Long Input(32) scenarios, containing 16k, 32k, and 32k irrelevant context tokens respectively, to measure whether the reasoning shift effect is reduced after fine-tuning. Notably, the model is trained only with 16k-token augmentations. 

\textbf{Results.} Our fine-tuned model demonstrates substantially improved robustness to irrelevant context in terms of reasoning length stability (Table~\ref{tab:multiturn_sft_math500}). Across all evaluated scenarios, the number of generated reasoning tokens produced by our model remains close to the corresponding baseline values where no irrelevant context is present. However, we note that the performance of our model with irrelevant context still degrades compared to the Baseline values. These results suggest that while robustly imitating high-level patterns might be trivial, maintaining the effectiveness of long reasoning traces under additional context pressure is a more challenging task.

\vspace{-10px}
\section{Discussion}\label{sect:discussion}
\vspace{-10px}

\textbf{Limitations}
We wish to highlight several important limitations of our work. First, while chosen on purpose to be easily controllable and easy to identify as redundant, our context conditions are relatively simple and synthetic: demonstrating and analyzing the described reasoning shift "in the wild" (using more realistic scenarios, including agentic ones) is of great interest for future work. Second, while we provide extensive empirical evidence that context conditions systematically shorten reasoning traces and suppress self-verification behaviors, our work does not offer a mechanistic account of how this effect arises within the model's internal representations. We leave a mechanistic investigation of this effect to future work, given the computational challenges of analyzing long inputs \citep{sharkey2025open} and the generalization limitations of interpretability techniques across context lengths and domains \citep{kramar2026building}. Lastly, we do not cover or develop any context management methods, including those based on recursive self-calls \citep{yang2025pencil, jin2025learningpromisescalinglanguage, ning2024skeletonofthought, zheng2025parallelr1, zhang2025recursive}.

In this paper, we find that different context conditions may affect the way reasoning LLMs tackle the same problems. In particular, we demonstrate that the distribution of high-level behavioral patterns, such as uncertainty management and self-verification, is fragile and can be suppressed by non-relevant context in the prompt. The efficiency of these self-checks is also reduced by irrelevant context. While for easier problems this may reduce overthinking, such behavioral shifts degrade performance on more challenging tasks. We also show that prompting has a limited effect on mitigating the described distributional shift and targeted fine-tuning helps to maintain high-level patterns of reasoning traces while having limited effect on the accuracy.

We hope our findings draw additional attention to both the robustness of reasoning models and the problem of context management for LLMs and LLM-based agents.

\bibliography{main}

\begin{thebibliography}{59}
\providecommand{\natexlab}[1]{#1}
\providecommand{\url}[1]{\texttt{#1}}
\expandafter\ifx\csname urlstyle\endcsname\relax
  \providecommand{\doi}[1]{doi: #1}\else
  \providecommand{\doi}{doi: \begingroup \urlstyle{rm}\Url}\fi

\bibitem[Agarwal et~al.(2025)Agarwal, Ahmad, Ai, Altman, Applebaum, Arbus, Arora, Bai, Baker, Bao, et~al.]{gpt-oss}
Sandhini Agarwal, Lama Ahmad, Jason Ai, Sam Altman, Andy Applebaum, Edwin Arbus, Rahul~K Arora, Yu~Bai, Bowen Baker, Haiming Bao, et~al.
\newblock gpt-oss-120b \& gpt-oss-20b model card.
\newblock \emph{arXiv preprint arXiv:2508.10925}, 2025.

\bibitem[Aggarwal et~al.(2025)Aggarwal, Kim, Lanchantin, Welleck, Weston, Kulikov, and Saha]{aggarwal2025optimalthinkingbench}
Pranjal Aggarwal, Seungone Kim, Jack Lanchantin, Sean Welleck, Jason Weston, Ilia Kulikov, and Swarnadeep Saha.
\newblock Optimalthinkingbench: Evaluating over and underthinking in llms.
\newblock \emph{arXiv preprint arXiv:2508.13141}, 2025.

\bibitem[{Anthropic}(2024)]{anthropic2024claude3}
{Anthropic}.
\newblock Introducing the next generation of claude, March 2024.
\newblock URL \url{https://www.anthropic.com/news/claude-3-family}.
\newblock Accessed: 2025-05-18.

\bibitem[{Anthropic}(2025)]{anthropic2025context_engineering}
{Anthropic}.
\newblock Effective context engineering for ai agents, 2025.
\newblock URL \url{https://www.anthropic.com/engineering/effective-context-engineering-for-ai-agents}.

\bibitem[Bogdan et~al.(2025)Bogdan, Macar, Nanda, and Conmy]{bogdan2025thought_anchors}
P.~C. Bogdan, U.~Macar, N.~Nanda, and A.~Conmy.
\newblock Thought anchors: Which {LLM} reasoning steps matter?
\newblock \emph{arXiv preprint arXiv:2506.19143}, jun 2025.
\newblock Preprint.

\bibitem[DeepSeek-AI et~al.(2025)DeepSeek-AI, Guo, Yang, Zhang, Song, Zhang, Xu, Zhu, Ma, Wang, and et~al.]{deepseek_r1}
DeepSeek-AI, Daya Guo, Dejian Yang, Haowei Zhang, Junxiao Song, Ruoyu Zhang, Runxin Xu, Qihao Zhu, Shirong Ma, Peiyi Wang, and Xiao~Bi et~al.
\newblock Deepseek-r1: Incentivizing reasoning capability in llms via reinforcement learning, 2025.
\newblock URL \url{https://arxiv.org/abs/2501.12948}.

\bibitem[Du et~al.(2025)Du, Tian, Ronanki, Rongali, Bodapati, Galstyan, Wells, Schwartz, Huerta, and Peng]{du2025context_length_alone}
Yufeng Du, Minyang Tian, Srikanth Ronanki, Subendhu Rongali, Sravan Bodapati, Aram Galstyan, Azton Wells, Roy Schwartz, Eliu~A Huerta, and Hao Peng.
\newblock Context length alone hurts llm performance despite perfect retrieval.
\newblock \emph{arXiv preprint arXiv:2510.05381}, 2025.

\bibitem[Dubey et~al.(2024)Dubey, Jauhri, Pandey, Kadian, Al-Dahle, Letman, Mathur, Schelten, Yang, Fan, et~al.]{dubey2024llama}
Abhimanyu Dubey, Abhinav Jauhri, Abhinav Pandey, Abhishek Kadian, Ahmad Al-Dahle, Aiesha Letman, Akhil Mathur, Alan Schelten, Amy Yang, Angela Fan, et~al.
\newblock The llama 3 herd of models.
\newblock \emph{arXiv preprint arXiv:2407.21783}, 2024.

\bibitem[Feng et~al.(2025)Feng, Kempe, Zhang, Jain, and Hartshorn]{feng2025characterizes_reasoning}
Yunzhen Feng, Julia Kempe, Cheng Zhang, Parag Jain, and Anthony Hartshorn.
\newblock What characterizes effective reasoning? revisiting length, review, and structure of cot.
\newblock \emph{arXiv preprint arXiv:2509.19284}, 2025.

\bibitem[Gandhi et~al.(2025)Gandhi, Chakravarthy, Singh, Lile, and Goodman]{gandhi2025cognitive_habits}
Kanishk Gandhi, Ayush Chakravarthy, Anikait Singh, Nathan Lile, and Noah~D Goodman.
\newblock Cognitive behaviors that enable self-improving reasoners, or, four habits of highly effective stars.
\newblock \emph{arXiv preprint arXiv:2503.01307}, 2025.

\bibitem[{Google DeepMind}(2025{\natexlab{a}})]{gemini3flash}
{Google DeepMind}.
\newblock Gemini 3 flash: frontier intelligence built for speed, 2025{\natexlab{a}}.
\newblock URL \url{https://blog.google/products-and-platforms/products/gemini/gemini-3-flash/}.

\bibitem[{Google DeepMind}(2025{\natexlab{b}})]{googledeepmind2025gemini25thinking}
{Google DeepMind}.
\newblock {Gemini 2.5: Our Newest Gemini Model with Thinking}.
\newblock \url{https://blog.google/technology/google-deepmind/gemini-model-thinking-updates-march-2025/#gemini-2-5-thinking}, 2025{\natexlab{b}}.
\newblock Accessed: 2025-04-07.

\bibitem[{Google DeepMind}(2026{\natexlab{a}})]{gemini31}
{Google DeepMind}.
\newblock Gemini 3.1 pro: A smarter model for your most complex tasks, 2026{\natexlab{a}}.
\newblock URL \url{https://blog.google/innovation-and-ai/models-and-research/gemini-models/gemini-3-1-pro/}.

\bibitem[{Google DeepMind}(2026{\natexlab{b}})]{gemma4}
{Google DeepMind}.
\newblock Gemma 4, 2026{\natexlab{b}}.
\newblock URL \url{https://deepmind.google/models/gemma/gemma-4/}.

\bibitem[Hassid et~al.(2025)Hassid, Synnaeve, Adi, and Schwartz]{hassid2025dont_overthink}
Michael Hassid, Gabriel Synnaeve, Yossi Adi, and Roy Schwartz.
\newblock Don't overthink it. preferring shorter thinking chains for improved llm reasoning.
\newblock \emph{arXiv preprint arXiv:2505.17813}, 2025.

\bibitem[Hsieh et~al.(2024)Hsieh, Sun, Kriman, Acharya, Rekesh, Jia, and Ginsburg]{hsieh2024ruler}
Cheng-Ping Hsieh, Simeng Sun, Samuel Kriman, Shantanu Acharya, Dima Rekesh, Fei Jia, and Boris Ginsburg.
\newblock {RULER}: What{\textquoteright}s the real context size of your long-context language models?
\newblock In \emph{First Conference on Language Modeling}, 2024.
\newblock URL \url{https://openreview.net/forum?id=kIoBbc76Sy}.

\bibitem[Jain et~al.(2024)Jain, Han, Gu, Li, Yan, Zhang, Wang, Solar-Lezama, Sen, and Stoica]{jain2024livecodebenchholisticcontaminationfree}
Naman Jain, King Han, Alex Gu, Wen-Ding Li, Fanjia Yan, Tianjun Zhang, Sida Wang, Armando Solar-Lezama, Koushik Sen, and Ion Stoica.
\newblock Livecodebench: Holistic and contamination free evaluation of large language models for code, 2024.
\newblock URL \url{https://arxiv.org/abs/2403.07974}.

\bibitem[Jin et~al.(2025)Jin, Cheng, Ankner, Saunshi, Elias, Yazdanbakhsh, Ragan-Kelley, Subramanian, and Carbin]{jin2025learningpromisescalinglanguage}
Tian Jin, Ellie~Y. Cheng, Zack Ankner, Nikunj Saunshi, Blake~M. Elias, Amir Yazdanbakhsh, Jonathan Ragan-Kelley, Suvinay Subramanian, and Michael Carbin.
\newblock Learning to keep a promise: Scaling language model decoding parallelism with learned asynchronous decoding, 2025.
\newblock URL \url{https://arxiv.org/abs/2502.11517}.

\bibitem[Kamradt(2023)]{needleinhaystack}
Greg Kamradt.
\newblock Needle in a haystack - pressure testing llms.
\newblock \url{https://github.com/gkamradt/LLMTest_NeedleInAHaystack}, 2023.

\bibitem[Karpathy(2015)]{char-rnn}
Andrej Karpathy.
\newblock char-rnn.
\newblock \url{https://github.com/karpathy/char-rnn}, 2015.

\bibitem[{Kimi Team}(2025)]{team2025kimik2}
{Kimi Team}.
\newblock Kimi k2: Open agentic intelligence.
\newblock \emph{arXiv preprint arXiv:2507.20534}, 2025.

\bibitem[Kojima et~al.(2022)Kojima, Gu, Reid, Matsuo, and Iwasawa]{zero_shot_cot_Kojima2022LargeLM}
Takeshi Kojima, Shixiang~Shane Gu, Machel Reid, Yutaka Matsuo, and Yusuke Iwasawa.
\newblock Large language models are zero-shot reasoners.
\newblock \emph{ArXiv}, abs/2205.11916, 2022.
\newblock URL \url{https://api.semanticscholar.org/CorpusID:249017743}.

\bibitem[Kram{\'a}r et~al.(2026)Kram{\'a}r, Engels, Wang, Chughtai, Shah, Nanda, and Conmy]{kramar2026building}
J{\'a}nos Kram{\'a}r, Joshua Engels, Zheng Wang, Bilal Chughtai, Rohin Shah, Neel Nanda, and Arthur Conmy.
\newblock Building production-ready probes for gemini.
\newblock \emph{arXiv preprint arXiv:2601.11516}, 2026.

\bibitem[Kumar et~al.(2024)Kumar, Zhuang, Agarwal, Su, Co-Reyes, Singh, Baumli, Iqbal, Bishop, Roelofs, et~al.]{kumar2024training_to_self_correct}
Aviral Kumar, Vincent Zhuang, Rishabh Agarwal, Yi~Su, John~D Co-Reyes, Avi Singh, Kate Baumli, Shariq Iqbal, Colton Bishop, Rebecca Roelofs, et~al.
\newblock Training language models to self-correct via reinforcement learning.
\newblock \emph{arXiv preprint arXiv:2409.12917}, 2024.

\bibitem[Kumaran et~al.(2026{\natexlab{a}})Kumaran, Conmy, Barbero, Osindero, Patraucean, and Velickovic]{kumaran2026how_llm_compute_confidence}
Dharshan Kumaran, Arthur Conmy, Federico Barbero, Simon Osindero, Viorica Patraucean, and Petar Velickovic.
\newblock How do llms compute verbal confidence.
\newblock \emph{arXiv preprint arXiv:2603.17839}, 2026{\natexlab{a}}.

\bibitem[Kumaran et~al.(2026{\natexlab{b}})Kumaran, Daw, Osindero, Velickovic, and Patraucean]{kumaran2026causal}
Dharshan Kumaran, Nathaniel Daw, Simon Osindero, Petar Velickovic, and Viorica Patraucean.
\newblock Causal evidence that language models use confidence to drive behavior.
\newblock \emph{arXiv preprint arXiv:2603.22161}, 2026{\natexlab{b}}.

\bibitem[Kumaran et~al.(2026{\natexlab{c}})Kumaran, Patraucean, Osindero, Velickovic, and Daw]{kumaran2026detect_own_errors}
Dharshan Kumaran, Viorica Patraucean, Simon Osindero, Petar Velickovic, and Nathaniel Daw.
\newblock How llms detect and correct their own errors: The role of internal confidence signals.
\newblock \emph{arXiv preprint arXiv:2604.22271}, 2026{\natexlab{c}}.

\bibitem[Laban et~al.(2026)Laban, Hayashi, Zhou, and Neville]{laban2026llms_get_lost}
Philippe Laban, Hiroaki Hayashi, Yingbo Zhou, and Jennifer Neville.
\newblock {LLM}s get lost in multi-turn conversation.
\newblock In \emph{The Fourteenth International Conference on Learning Representations}, 2026.
\newblock URL \url{https://openreview.net/forum?id=VKGTGGcwl6}.

\bibitem[Li et~al.(2024)Li, Zhang, Do, Yue, and Chen]{li2024long_icl_bench}
Tianle Li, Ge~Zhang, Quy~Duc Do, Xiang Yue, and Wenhu Chen.
\newblock Long-context llms struggle with long in-context learning.
\newblock \emph{arXiv preprint arXiv:2404.02060}, 2024.

\bibitem[Liang et~al.(2025)Liang, Song, Li, Yang, Zhang, Mi, and Yu]{liang2025towards}
Zhenwen Liang, Linfeng Song, Yang Li, Tao Yang, Feng Zhang, Haitao Mi, and Dong Yu.
\newblock Towards solving more challenging imo problems via decoupled reasoning and proving.
\newblock \emph{arXiv preprint arXiv:2507.06804}, 2025.

\bibitem[Lightman et~al.(2024)Lightman, Kosaraju, Burda, Edwards, Baker, Lee, Leike, Schulman, Sutskever, and Cobbe]{lightman2023lets_verify_step_by_step}
Hunter Lightman, Vineet Kosaraju, Yuri Burda, Harrison Edwards, Bowen Baker, Teddy Lee, Jan Leike, John Schulman, Ilya Sutskever, and Karl Cobbe.
\newblock Let's verify step by step.
\newblock In \emph{The twelfth international conference on learning representations}, 2024.

\bibitem[Ling et~al.(2025)Ling, Liu, Yan, Yang, Lin, Fan, Shen, Du, and Chen]{ling2025longreason}
Zhan Ling, Kang Liu, Kai Yan, Yifan Yang, Weijian Lin, Ting-Han Fan, Lingfeng Shen, Zhengyin Du, and Jiecao Chen.
\newblock Longreason: A synthetic long-context reasoning benchmark via context expansion.
\newblock \emph{arXiv preprint arXiv:2501.15089}, 2025.

\bibitem[Liu et~al.(2025{\natexlab{a}})Liu, Mei, Lin, Xue, Wang, Xu, Wu, Zhang, Lin, Dong, et~al.]{liu2025deepseekv3_2}
Aixin Liu, Aoxue Mei, Bangcai Lin, Bing Xue, Bingxuan Wang, Bingzheng Xu, Bochao Wu, Bowei Zhang, Chaofan Lin, Chen Dong, et~al.
\newblock Deepseek-v3. 2: Pushing the frontier of open large language models.
\newblock \emph{arXiv preprint arXiv:2512.02556}, 2025{\natexlab{a}}.

\bibitem[Liu et~al.(2025{\natexlab{b}})Liu, Zhu, Bai, He, Liao, Que, Wang, Zhang, Zhang, Zhang, et~al.]{liu2025long_context_survey}
Jiaheng Liu, Dawei Zhu, Zhiqi Bai, Yancheng He, Huanxuan Liao, Haoran Que, Zekun Wang, Chenchen Zhang, Ge~Zhang, Jiebin Zhang, et~al.
\newblock A comprehensive survey on long context language modeling.
\newblock \emph{arXiv preprint arXiv:2503.17407}, 2025{\natexlab{b}}.

\bibitem[Luong et~al.(2025)Luong, Hwang, Nguyen, Ghiasi, Chervonyi, Seo, Kim, Bingham, Lee, Mishra, et~al.]{imobench}
Minh-Thang Luong, Dawsen Hwang, Hoang~H Nguyen, Golnaz Ghiasi, Yuri Chervonyi, Insuk Seo, Junsu Kim, Garrett Bingham, Jonathan Lee, Swaroop Mishra, et~al.
\newblock Towards robust mathematical reasoning.
\newblock In \emph{Proceedings of the 2025 Conference on Empirical Methods in Natural Language Processing}, pages 35406--35430, 2025.

\bibitem[{Moonshot AI}(2025)]{kimik2thinking}
{Moonshot AI}.
\newblock Introducing kimi k2 thinking, 2025.
\newblock URL \url{https://www.kimi.com/blog/kimi-k2-thinking}.

\bibitem[Muennighoff et~al.(2025)Muennighoff, Yang, Shi, Li, Fei-Fei, Hajishirzi, Zettlemoyer, Liang, Cand{\`e}s, and Hashimoto]{muennighoff2025s1}
Niklas Muennighoff, Zitong Yang, Weijia Shi, Xiang~Lisa Li, Li~Fei-Fei, Hannaneh Hajishirzi, Luke Zettlemoyer, Percy Liang, Emmanuel Cand{\`e}s, and Tatsunori Hashimoto.
\newblock s1: Simple test-time scaling.
\newblock \emph{arXiv preprint arXiv:2501.19393}, 2025.

\bibitem[Ning et~al.(2024)Ning, Lin, Zhou, Wang, Yang, and Wang]{ning2024skeletonofthought}
Xuefei Ning, Zinan Lin, Zixuan Zhou, Zifu Wang, Huazhong Yang, and Yu~Wang.
\newblock Skeleton-of-thought: Prompting {LLM}s for efficient parallel generation.
\newblock In \emph{The Twelfth International Conference on Learning Representations}, 2024.
\newblock URL \url{https://openreview.net/forum?id=mqVgBbNCm9}.

\bibitem[OpenAI(2024)]{openai_o1}
OpenAI.
\newblock Openai o1 system card, 2024.
\newblock URL \url{https://arxiv.org/abs/2412.16720}.

\bibitem[{QwenTeam}(2026)]{qwen35blog}
{QwenTeam}.
\newblock Qwen3.5: Towards native multimodal agents, February 2026.
\newblock URL \url{https://qwen.ai/blog?id=qwen3.5}.

\bibitem[Rein et~al.(2023)Rein, Hou, Stickland, Petty, Pang, Dirani, Michael, and Bowman]{rein2023gpqagraduatelevelgoogleproofqa}
David Rein, Betty~Li Hou, Asa~Cooper Stickland, Jackson Petty, Richard~Yuanzhe Pang, Julien Dirani, Julian Michael, and Samuel~R. Bowman.
\newblock Gpqa: A graduate-level google-proof q\&a benchmark, 2023.
\newblock URL \url{https://arxiv.org/abs/2311.12022}.

\bibitem[Shao et~al.(2024)Shao, Wang, Zhu, Xu, Song, Bi, Zhang, Zhang, Li, Wu, et~al.]{shao2024deepseekmath}
Zhihong Shao, Peiyi Wang, Qihao Zhu, Runxin Xu, Junxiao Song, Xiao Bi, Haowei Zhang, Mingchuan Zhang, YK~Li, Yang Wu, et~al.
\newblock Deepseekmath: Pushing the limits of mathematical reasoning in open language models.
\newblock \emph{arXiv preprint arXiv:2402.03300}, 2024.

\bibitem[Shao et~al.(2025)Shao, Luo, Lu, Ren, Hu, Ye, Gou, Ma, and Zhang]{shao2025deepseekmathv2}
Zhihong Shao, Yuxiang Luo, Chengda Lu, ZZ~Ren, Jiewen Hu, Tian Ye, Zhibin Gou, Shirong Ma, and Xiaokang Zhang.
\newblock Deepseekmath-v2: Towards self-verifiable mathematical reasoning.
\newblock \emph{arXiv preprint arXiv:2511.22570}, 2025.

\bibitem[Sharkey et~al.(2025)Sharkey, Chughtai, Batson, Lindsey, Wu, Bushnaq, Goldowsky-Dill, Heimersheim, Ortega, Bloom, et~al.]{sharkey2025open}
Lee Sharkey, Bilal Chughtai, Joshua Batson, Jack Lindsey, Jeff Wu, Lucius Bushnaq, Nicholas Goldowsky-Dill, Stefan Heimersheim, Alejandro Ortega, Joseph Bloom, et~al.
\newblock Open problems in mechanistic interpretability.
\newblock \emph{arXiv preprint arXiv:2501.16496}, 2025.

\bibitem[Su et~al.(2025)Su, Healey, Nakov, and Cardie]{su2025between_underthinking}
Jinyan Su, Jennifer Healey, Preslav Nakov, and Claire Cardie.
\newblock Between underthinking and overthinking: An empirical study of reasoning length and correctness in llms.
\newblock \emph{arXiv preprint arXiv:2505.00127}, 2025.

\bibitem[{Team Olmo}(2025)]{olmo3}
{Team Olmo}.
\newblock Olmo 3.
\newblock \emph{arXiv preprint arXiv:2512.13961}, 2025.

\bibitem[Tekparmak and Ömer Kaya(2025)]{memagent}
Atakan Tekparmak and Ömer Kaya.
\newblock mem-agent: Equipping llm agents with memory using rl, October 2025.
\newblock URL \url{https://huggingface.co/blog/driaforall/mem-agent-blog}.
\newblock Community Article.

\bibitem[Venhoff et~al.(2025)Venhoff, Arcuschin, Torr, Conmy, and Nanda]{venhoff2025understanding}
C.~Venhoff, I.~Arcuschin, P.~Torr, A.~Conmy, and N.~Nanda.
\newblock Understanding reasoning in thinking language models via steering vectors.
\newblock In \emph{Workshop on Reasoning and Planning for Large Language Models at ICLR 2025}, 2025.
\newblock URL \url{https://arxiv.org/abs/2506.18167}.

\bibitem[Wang et~al.(2024)Wang, Ma, Zhang, Ni, Chandra, Guo, Ren, Arulraj, He, Jiang, et~al.]{wang2024mmlu}
Yubo Wang, Xueguang Ma, Ge~Zhang, Yuansheng Ni, Abhranil Chandra, Shiguang Guo, Weiming Ren, Aaran Arulraj, Xuan He, Ziyan Jiang, et~al.
\newblock Mmlu-pro: A more robust and challenging multi-task language understanding benchmark.
\newblock \emph{Advances in Neural Information Processing Systems}, 37:\penalty0 95266--95290, 2024.

\bibitem[Wei et~al.(2022)Wei, Wang, Schuurmans, Bosma, ichter, Xia, Chi, Le, and Zhou]{cot_wei_2022}
Jason Wei, Xuezhi Wang, Dale Schuurmans, Maarten Bosma, brian ichter, Fei Xia, Ed~Chi, Quoc~V Le, and Denny Zhou.
\newblock Chain-of-thought prompting elicits reasoning in large language models.
\newblock In S.~Koyejo, S.~Mohamed, A.~Agarwal, D.~Belgrave, K.~Cho, and A.~Oh, editors, \emph{Advances in Neural Information Processing Systems}, volume~35, pages 24824--24837. Curran Associates, Inc., 2022.
\newblock \doi{10.52202/068431-1800}.
\newblock URL \url{https://proceedings.neurips.cc/paper_files/paper/2022/file/9d5609613524ecf4f15af0f7b31abca4-Paper-Conference.pdf}.

\bibitem[Wu et~al.(2025)Wu, Wang, Ye, Du, Jegelka, and Wang]{wu2025understand_CoT_lenth}
Yuyang Wu, Yifei Wang, Ziyu Ye, Tianqi Du, Stefanie Jegelka, and Yisen Wang.
\newblock When more is less: Understanding chain-of-thought length in llms.
\newblock \emph{arXiv preprint arXiv:2502.07266}, 2025.

\bibitem[Xu et~al.(2026)Xu, Lin, Xue, Wang, Xu, Wu, Zhang, Lin, Dong, Ling, et~al.]{deepseekv42026}
Anyi Xu, Bangcai Lin, Bing Xue, Bingxuan Wang, Bingzheng Xu, Bochao Wu, Bowei Zhang, Chaofan Lin, Chen Dong, Chenchen Ling, et~al.
\newblock Deepseek-v4: Towards highly efficient million-token context intelligence.
\newblock \emph{arXiv preprint arXiv:2606.19348}, 2026.

\bibitem[Yan et~al.(2025)Yan, Shen, Liu, Jiang, Zhang, Shao, and Zhuang]{yan2025inftythink}
Yuchen Yan, Yongliang Shen, Yang Liu, Jin Jiang, Mengdi Zhang, Jian Shao, and Yueting Zhuang.
\newblock Inftythink: Breaking the length limits of long-context reasoning in large language models.
\newblock \emph{arXiv preprint arXiv:2503.06692}, 2025.

\bibitem[Yang et~al.(2025{\natexlab{a}})Yang, Li, Yang, Zhang, Hui, Zheng, Yu, Gao, Huang, Lv, Zheng, Liu, Zhou, Huang, Hu, Ge, Wei, Lin, Tang, Yang, Tu, Zhang, Yang, Yang, Zhou, Zhou, Lin, Dang, Bao, Yang, Yu, Deng, Li, Xue, Li, Zhang, Wang, Zhu, Men, Gao, Liu, Luo, Li, Tang, Yin, Ren, Wang, Zhang, Ren, Fan, Su, Zhang, Zhang, Wan, Liu, Wang, Cui, Zhang, Zhou, and Qiu]{qwen3}
An~Yang, Anfeng Li, Baosong Yang, Beichen Zhang, Binyuan Hui, Bo~Zheng, Bowen Yu, Chang Gao, Chengen Huang, Chenxu Lv, Chujie Zheng, Dayiheng Liu, Fan Zhou, Fei Huang, Feng Hu, Hao Ge, Haoran Wei, Huan Lin, Jialong Tang, Jian Yang, Jianhong Tu, Jianwei Zhang, Jianxin Yang, Jiaxi Yang, Jing Zhou, Jingren Zhou, Junyang Lin, Kai Dang, Keqin Bao, Kexin Yang, Le~Yu, Lianghao Deng, Mei Li, Mingfeng Xue, Mingze Li, Pei Zhang, Peng Wang, Qin Zhu, Rui Men, Ruize Gao, Shixuan Liu, Shuang Luo, Tianhao Li, Tianyi Tang, Wenbiao Yin, Xingzhang Ren, Xinyu Wang, Xinyu Zhang, Xuancheng Ren, Yang Fan, Yang Su, Yichang Zhang, Yinger Zhang, Yu~Wan, Yuqiong Liu, Zekun Wang, Zeyu Cui, Zhenru Zhang, Zhipeng Zhou, and Zihan Qiu.
\newblock Qwen3 technical report, 2025{\natexlab{a}}.
\newblock URL \url{https://arxiv.org/abs/2505.09388}.

\bibitem[Yang et~al.(2025{\natexlab{b}})Yang, Srebro, McAllester, and Li]{yang2025pencil}
Chenxiao Yang, Nathan Srebro, David McAllester, and Zhiyuan Li.
\newblock Pencil: Long thoughts with short memory.
\newblock \emph{arXiv preprint arXiv:2503.14337}, 2025{\natexlab{b}}.

\bibitem[Ye et~al.(2025)Ye, Huang, Xiao, Chern, Xia, and Liu]{ye2025limoreasoning}
Yixin Ye, Zhen Huang, Yang Xiao, Ethan Chern, Shijie Xia, and Pengfei Liu.
\newblock Limo: Less is more for reasoning, 2025.
\newblock URL \url{https://arxiv.org/abs/2502.03387}.

\bibitem[Yu et~al.(2025)Yu, Zhang, Zhu, Yuan, Zuo, Yue, Dai, Fan, Liu, Liu, et~al.]{yu2025dapo}
Qiying Yu, Zheng Zhang, Ruofei Zhu, Yufeng Yuan, Xiaochen Zuo, Yu~Yue, Weinan Dai, Tiantian Fan, Gaohong Liu, Lingjun Liu, et~al.
\newblock Dapo: An open-source llm reinforcement learning system at scale.
\newblock \emph{arXiv preprint arXiv:2503.14476}, 2025.

\bibitem[Zhang et~al.(2025)Zhang, Kraska, and Khattab]{zhang2025recursive}
Alex~L Zhang, Tim Kraska, and Omar Khattab.
\newblock Recursive language models.
\newblock \emph{arXiv preprint arXiv:2512.24601}, 2025.

\bibitem[Zheng et~al.(2025)Zheng, Zhang, Yu, Wang, Dai, Liu, Bao, Huang, Huang, and Yu]{zheng2025parallelr1}
Tong Zheng, Hongming Zhang, Wenhao Yu, Xiaoyang Wang, Runpeng Dai, Rui Liu, Huiwen Bao, Chengsong Huang, Heng Huang, and Dong Yu.
\newblock Parallel-r1: Towards parallel thinking via reinforcement learning, 2025.
\newblock URL \url{https://arxiv.org/abs/2509.07980}.

\end{thebibliography}
\bibliographystyle{plainnat}


\newpage
\appendix
\newpage
\appendix
\section{The Main Experiment details}\label{app:exp_details}

\subsection{Prompt details}

\begin{tcolorbox}[colback=gray!5, colframe=gray!75, title=\textbf{Baseline}, sharp corners, breakable]

    \begin{quote}
    Please reason step-by-step and put the final answer within \textbackslash boxed\{\}.
    \newline
    \newline
    <Problem>
    \end{quote}

\end{tcolorbox}

\begin{tcolorbox}[colback=gray!5, colframe=gray!75, title=\textbf{Subtask}, sharp corners, breakable]

    \begin{quote}
    You will be given two independent problems, solve them separately. For each problem please reason step by step, and put your final answer within \textbackslash boxed\{\}.
    \newline
    \newline
    First Problem:
    \newline
    <Problem>
    \newline
    \newline
    Second Problem:
    \newline
    \newline
    <Problem>
    \end{quote}

\end{tcolorbox}

\begin{tcolorbox}[colback=gray!5, colframe=gray!75, title=\textbf{Long Input}, sharp corners, breakable]

    \begin{quote}
    \# Old data:
    \newline
    \newline
    <Shakespeare's plays>
    \newline
    \newline
    Old data ends.
    \newline
    \newline
    Please reason step-by-step and put the final answer within \textbackslash boxed\{\}.
    \newline
    \newline
    <Problem>
    \end{quote}

\end{tcolorbox}

For Long Input setup, we insert 64000 tokens of Shakespeare's plays from the \citep{char-rnn} to the prompt. We use unrelated text as a simple control to ensure that the model can easily recognize the additional context as irrelevant and do not spend their effort investigating the attached prefix. 

\begin{tcolorbox}[colback=gray!5, colframe=gray!75, title=\textbf{Baseline, max reasoning effort}, sharp corners, breakable]

    \begin{quote}
    Reasoning Effort: Absolute maximum with no shortcuts permitted.
You MUST be very thorough in your thinking and comprehensively decompose the
problem to resolve the root cause, rigorously stress-testing your logic against all potential
paths, edge cases, and adversarial scenarios.
Explicitly write out your entire deliberation process, documenting every intermediate
step, considered alternative, and rejected hypothesis to ensure absolutely no assumption
is left unchecked. Put the final answer within \textbackslash boxed\{\}.
    \newline
    \newline
    <Problem>
    \end{quote}

\end{tcolorbox}

\subsection{Inference details}

We use OpenRouter API for running the main experiments. To make sure that the results across setups are compatible, we use a single fixed vendor for each model. Here we list the providers and the sampling parameters used in our experiments:

\begin{itemize}
    \item Qwen3.5-27B: Alibaba Cloud Int.;  
    \item Gemma 4 31B: Parasail;
    \item gpt-oss-120b: Together AI; reasoning effort xhigh.
    \item Gemini 3 Flash Preview: Google AI Studio;
    \item Kimi K2 Thinking: Google Vertex;
\end{itemize}

For each model, we use default sampling parameters and thinking budget of 80,000 tokens across all context conditions. This compute budget is enough for most setups/models, except rare (less than 2\% of generations) cases for Kimi K2 Thinking, we exclude unfinished generations from our analysis, keeping all side-by-side comparisons over the subset of problems where generations were finished naturally by the model.

We report the amount of reasoning tokens reported by the providers, which matches our estimations with manual tokenization where applicable.

For Olmo3 experiments, we use local inference with transformers library, using recommended sampling  parameters: temperature 0.6, top\_p: 0.95, max new tokens 16384.

\subsection{Compute resources} \label{app:compute}

Our main evaluation compute was spent on API-based inference for IMO AnswerBench, GPQA-Diamond, and LiveCodeBench. We accessed the evaluated models through OpenRouter. Across the main experiments, typical reasoning traces contain roughly $5$k--$30$k generated tokens per example, depending on the benchmark, model, and prompting condition. Table~\ref{tab:api_compute_costs} summarizes the API pricing used for cost estimation. As a concrete reference point, a full GPQA-Diamond run for the evaluated model set cost approximately $\$180$; the total API cost of the main benchmark evaluations is therefore on the order of several hundred dollars.

A second source of compute was used for the analysis experiments, including the analysis and OLMo evaluations. These experiments were run on local GPU infrastructure. Since these runs include exploratory analyses, failed runs, and intermediate ablations that are not all reported in the paper, we report a conservative upper-bound estimate rather than an exact accounting. Overall, we estimate that these analysis and OLMo evaluation experiments required at most $1000$ NVIDIA A100 GPU-hours.

\begin{table}[h]
\centering
\small
\setlength{\tabcolsep}{5pt}
\renewcommand{\arraystretch}{1.1}

\begin{tabular}{lcc}
\toprule
\textbf{Model} & \textbf{Input} & \textbf{Output} \\
\midrule
Qwen3.5-27B & \$0.195 / M & \$1.56 / M \\
Gemma 4 31B & \$0.14 / M & \$0.40 / M \\
gpt-oss-120b & \$0.039 / M & \$0.18 / M \\
Gemini 3 Flash Preview & \$0.50 / M & \$3.00 / M \\
Kimi K2 Thinking & \$0.60 / M & \$2.50 / M \\
\bottomrule
\end{tabular}
\caption{API pricing used for estimating evaluation costs. Prices are reported in USD per million tokens, following the OpenRouter model pages at the time of the experiments.}
\label{tab:api_compute_costs}
\end{table}

\section{Setup design}\label{app:setup_design}

In this section, we discuss our motivation behind each of the three setups we consider for our main experiments. Our central research question is: “Can a model solve an isolated subproblem as effectively when surrounded by irrelevant context as it does in isolation?” Although one may ask why a model should operate with irrelevant context when a clean context is available, we believe each condition is worth investigating:

Subtask: In reasoning-intensive tasks, models may encounter independent subtasks that can be stated and solved without the preceding derivation. For example, solving an equation may not require the geometric context from which it was derived. Reproducing such situations exactly is difficult, so combining independent problems from the same domain provides a reasonable controlled approximation.

Multi-turn: Our setup complements \cite{laban2026llms_get_lost}: they study related, underspecified conversations where models make early assumptions and fail to recover, whereas we test whether a self-contained problem is solved differently merely because it follows unrelated interactions. Although synthetic, this setup isolates reasoning shifts without confounding them with misleading assumptions or corrections.

Long Input: Long-running scientific and coding agents often accumulate documentation, tool outputs, intermediate results, and previous attempts that are not needed for the current subproblem. We deliberately use Shakespeare as a simple, conservative condition in which the additional text is clearly irrelevant. This minimizes the confound that a model might interpret a distractor as useful and spend additional reasoning trying to incorporate it.

\section{Long Input with relevant context}\label{app:long_input_relevant}

We additionally evaluated two models, gpt-oss-120b and Qwen3.5-27B, on IMOAnswerBench under the long-input condition, replacing the irrelevant context with IMO lemmas and their proofs from \cite{liang2025towards}. The results are presented in Table \ref{tab:long_context_imo_lemmas} and show a similar trend: both models generate fewer thinking tokens and achieve lower accuracy under long input, even when the context consists of domain-relevant mathematical content.  

\begin{table}[ht]

    \centering
    \renewcommand{\arraystretch}{1.15}
    \setlength{\tabcolsep}{10pt}

    \begin{tabular}{l cc cc cc}
        \toprule
        & \multicolumn{2}{c}{Baseline}
        & \multicolumn{2}{c}{\makecell{Long Input\\(Shakespeare)}}
        & \multicolumn{2}{c}{\makecell{Long Input\\(IMO lemmas)}} \\
        \cmidrule(lr){2-3}
        \cmidrule(lr){4-5}
        \cmidrule(lr){6-7}
        Model & Acc. & Tokens & Acc. & Tokens & Acc. & Tokens \\
        \midrule
        gpt-oss-120b
            & 73.8 & 24,180
            & \cellcolor{red!13}64.0
            & \cellcolor{red!51}11,876
            & \cellcolor{red!14}63.5
            & \cellcolor{red!54}11,055 \\
        Qwen3.5-27B
            & 74.5 & 28,771
            & \cellcolor{red!9}67.8
            & \cellcolor{red!43}16,415
            & \cellcolor{red!9}68.0
            & \cellcolor{red!46}15,465 \\
        \bottomrule
    \end{tabular}
\caption{Model performance on IMOAnswerBench under Long Input scenario with different content used as prefix.}
\label{tab:long_context_imo_lemmas}
\end{table}

We note that we do not claim that these particular lemmas are directly relevant or useful for solving the problem in the query. Rather, they provide domain-relevant mathematical context. Determining whether context containing information directly useful for solving each specific problem produces the same effect would require a different experimental setup and is beyond the scope of our main research question, which is stated as follows: “Can a model solve an isolated subproblem as effectively when surrounded by irrelevant context as it does in isolation?”

\section{Additional experiment on code generation task}\label{app:lcb_code_experiment}
To further demonstrate that the observed reduction in reasoning length generalizes beyond mathematical reasoning tasks, we additionally evaluate reasoning models on code generation problems from the \textit{LiveCodeBench} dataset \citep{jain2024livecodebenchholisticcontaminationfree}. Specifically, we use the \texttt{code\_generation\_lite} \texttt{v6} subset, which contains 1055 competitive programming and algorithmic code generation tasks.

For Qwen3.5-27B, we use a maximum generation budget of 80k tokens together with a 32k-token input limit. For gpt-oss-120b, we use a 100k-token generation budget and a 28k-token input limit.

\begin{table}[h]
\centering

\begin{tabular}{lcccccccccc}
\toprule
\multirow{2}{*}{\textbf{Model}} 
& \multicolumn{2}{c}{\textbf{Baseline}} 
& \multicolumn{3}{c}{\textbf{Subtask}} 
& \multicolumn{2}{c}{\textbf{Long Input}} 
& \multicolumn{2}{c}{\textbf{Multi-turn}} \\
\cmidrule(lr){2-3} 
\cmidrule(lr){4-6} 
\cmidrule(lr){7-8} 
\cmidrule(lr){9-10}
& \textbf{Acc.} 
& \textbf{Tok.} 
& $\textbf{Acc}_1$ 
& $\textbf{Acc}_2$
& \textbf{Tok.} 
& \textbf{Acc.} 
& \textbf{Tok.} 
& \textbf{Acc.} 
& \textbf{Tok.} \\
\midrule
Qwen3.5-27B 
& 87 
& 22,837 
& \cellcolor{red!10}78 
& \cellcolor{red!12}78 
& \cellcolor{red!30}16,302 
& \cellcolor{red!2}85 
& \cellcolor{red!26}17,429 
& \cellcolor{red!8}81 
& \cellcolor{red!50}6,222 \\

gpt-oss-120b 
& 89 
& 10,666 
& \cellcolor{red!20}51
& \cellcolor{red!15}86 
& \cellcolor{red!23}7,675 
& \cellcolor{red!5}88 
& \cellcolor{red!26}7,173 
& \cellcolor{red!8}83 
& \cellcolor{red!30}6,156 \\
\bottomrule
\end{tabular}
\caption{Model performance with accuracy and average number of generated reasoning tokens on LiveCodeBench. For the Subtask setup, accuracies are reported separately for the two generated subtasks. Background color represents the \textcolor{red!80}{relative change} from the baseline values.}
\label{tab:lcb_code}
\end{table}

Here, we observe effects similar to those reported in Table~\ref{tab:main_exp}. Both Qwen3.5-27B and gpt-oss-120b exhibit substantially shorter reasoning traces when irrelevant or additional contextual information is introduced into the prompt. In particular, Qwen3.5-27B demonstrates a dramatic reduction in reasoning length in the multi-turn setting, generating nearly $3.6\times$ fewer reasoning tokens compared to the baseline setup.

Interestingly, gpt-oss-120b also exhibits a noticeable degradation in performance in the Subtask setting, where the model is required to solve two tasks within a single response.

\section{Subtask evaluation details}\label{app:subtask_evaluation_details}

To evaluate average performance in the \textbf{Subtask} scenario, we use two separate judge calls, evaluating the correctness of each subproblem. Results are presented in Table \ref{tab:subproblems}. We note that each problem appears once as a first subtask and once as a second subtask.

\begin{table}[h]
\centering

\begin{tabular}{lccc}
\toprule
\textbf{Model} & \textbf{Baseline} & \textbf{First subproblem} & \textbf{Second subproblem} \\
\midrule
Qwen3.5-27B & 74.5 & \cellcolor{red!10}66.8 & \cellcolor{red!22}58.0 \\
gpt-oss-120b & 73.8 & \cellcolor{red!14}63.8 & \cellcolor{red!13}64.3 \\
Gemini 3 Flash Preview & 82.8 & \cellcolor{red!18}68.3 & \cellcolor{red!21}65.8 \\
Kimi K2 Thinking & 74.8 & \cellcolor{red!9}68.0 & \cellcolor{red!17}62.0 \\
\bottomrule
\end{tabular}
\caption{Model accuracy on each subproblem in the Subproblem setup. Background color represents the \textcolor{red!80}{relative change} from the baseline values.}
\label{tab:subproblems}
\end{table}


\section{Analysis experiments additional details}\label{app:exp_details_analysis}

\subsection{Verbalized confidence experiment}
\label{sec:app_verbalized_confidence}

\begin{table}[h]
\centering

\begin{tabular}{lcccc}
\toprule
\textbf{Trace} & \textbf{Baseline} & \textbf{Long Input(128)} & \textbf{Long Input(16k)} & \textbf{Multi-turn} \\
\midrule
\textbf{Baseline}       & 85.9\%     & 88.4\%          & 90.4\%           & 88.8\% \\
\textbf{Long Input}   & 82.1\%     & 87.1\%          & 90.2\%           & 86.7\% \\
\textbf{Multi-turn}   & 83.0\%     & 86.5\%           & 89.0\%           & 86.7\% \\
\midrule
\end{tabular}

\caption{Verbalized confidence experiment with last 64 tokens removed, using numeric scores. The same reasoning traces receive higher self-confidence scores under non-baseline context conditions. Table represents the ratio of traces with the highest self-confidence scores. Rows represent how traces were generated, columns represent the context used during self-confidence evaluation. Qwen3-32B, MATH-500.}
\label{tab:self_confidence_numeric}

\end{table}

To generate self-confidence scores, we reuse the prompt from \cite{kumaran2026how_llm_compute_confidence}. However, since we aim to estimate the confidence in unfinished reasoning trace, we inject the prompt as an intervention into the reasoning process, not as next turn of the conversation.

\begin{tcolorbox}[colback=gray!5, colframe=gray!75, title=\textbf{Prompt used to generate self-confidence scores}, sharp corners, breakable]

    \begin{quote}
    ...Wait, let me estimate the confidence in my partial solution.
\newline
I will classify my confidence into one of the following classes based on how likely my answer is to be correct (NO REASONING OR EXPLANATION):
\newline
\newline
- "No chance" (0.0–0.1)\newline
- "Really unlikely" (0.1–0.2)\newline
- "Chances are slight" (0.2–0.3)\newline
- "Unlikely" (0.3–0.4)\newline
- "Less than even" (0.4–0.5)\newline
- "Better than even" (0.5–0.6)\newline
- "Likely" (0.6–0.7)\newline
- "Very good chance" (0.7–0.8)\newline
- "Highly likely" (0.8–0.9)\newline
- "Almost certain" (0.9–1.0)\newline
\newline
\newline
Each category reflects the probability that the answer is correct.
\newline
I will format my confidence as
**Confidence**: CLASS
where CLASS is one of the names above (without the probability ranges).

Okay, my confidence is **Confidence**:
    \end{quote}

\end{tcolorbox}

To obtain more robust results, we also tried an alternative prompt where we ask the model to provide a numerical score from 0 to 99. We observe that the model typically selects between the two highest confidence classes (i.e., ''Highly likely'' and ''Almost certain'') in the categorical prompt, and between 98 and 99 in the numerical variant. Thus, we report the ratios of the maximum confidence scores. We did not perform calibration because our analysis focuses on statistical changes in confidence between different context conditions rather than on the absolute accuracy of the reported probabilities. Results for numeric evaluation are presented in Table \ref{tab:self_confidence_numeric}. We observe a similar trend: for all reasoning traces, the proportion of problems assigned the maximum confidence score (99) was significantly lower under the baseline context condition than under each non-baseline condition; with $p < 0.001$ one-sided McNemar tests on paired per-task observations.

We additionally evaluate the confidence scores from shorter reasoning prefixes, keeping 50\% of the reasoning trace before making a similar intervention. This allows us to assess whether context-induced confidence inflation manifests early in the reasoning process or only emerges after the model has had sufficient tokens to develop its solution. Results are presented in Table \ref{tab:self_confidence_middle}.

\begin{table}[b]
\centering

\begin{tabular}{lcccc}
\toprule
\textbf{Trace} & \textbf{Baseline} & \textbf{Long Input(128)} & \textbf{Long Input(16k)} & \textbf{Multi-turn} \\
\midrule
\textbf{Baseline}       & 38.0\%     & 50.3\%          & 57.8\%           & 48.6\% \\
\textbf{Long Input}   & 16.8\%     & 28.3\%          & 36.8\%           & 26.8\% \\
\textbf{Multi-turn}   & 22.5\%     & 33.7\%           & 44.8\%           & 31.8\% \\
\midrule
\end{tabular}
\caption{Verbalized confidence experiment with the first half of the trace being evaluated. The same reasoning traces receive higher self-confidence scores under non-baseline context conditions. Table represents the ratio of traces with the highest self-confidence scores. Rows represent how traces were generated, columns represent the context used during self-confidence evaluation. Qwen3-32B, MATH-500.}
\label{tab:self_confidence_middle}
\end{table}

\begin{table}[]
\centering

\begin{tabular}{ll ccc}
\toprule
\textbf{Model} & & \textbf{Baseline} & \textbf{Long Input} & \textbf{Multi-turn} \\
\midrule
\multirow{2}{*}{Qwen3-32B}
 & Total reasoning tokens           & 3,421 & 2,375 & 2,825 \\
 & First candidate answer position  & 797  & 698  & 753  \\
\midrule
\multirow{2}{*}{Qwen3.5-27B}
 & Total reasoning tokens           & 4,689 & 2,545 & 3,017 \\
 & First candidate answer position  & 574  & 516  & 533  \\
\midrule
\multirow{2}{*}{Gemma-4-31B}
 & Total reasoning tokens           & 1,929 & 1,227 & 1,391 \\
 & First candidate answer position  & 656  & 638  & 649  \\
\bottomrule
\end{tabular}
\caption{Average reasoning length and first candidate answer position across models and settings, MATH-500.}
\label{tab:first_answer_position}
\end{table}

\begin{table}[]
\centering

\begin{tabular}{ll cccc}

\toprule
\textbf{Model} & & \textbf{Baseline} & \textbf{Long Input} & \textbf{Multi-turn} & \textbf{Multi-turn(4)} \\
\midrule
\multirow{3}{*}{Qwen3.5-27B}
 & Baseline    & 12.2 & 14.8 & 18.8 & 28.8 \\
 & Long Input  &  3.1 & 11.0 & 11.3 & 14.1 \\
 & Multi-turn &  5.3 &  5.7 & 11.1 & 20.1 \\
\midrule
\multirow{3}{*}{Gemma-4-31B}
 & Baseline    & 35.0 & 35.5 & 56.8 & 60.7 \\
 & Long Input  & 16.1 & 16.1 & 32.9 & 34.2 \\
 & Multi-turn & 16.9 & 20.7 & 33.5 & 39.3 \\
\bottomrule
\end{tabular}
\caption{The ratio of finished samples during resampling experiment. Rows represent the source of the trace, column - the context condition during resampling. The same almost finished reasoning traces tend to finish earlier under non-baseline conditions. Multi-turn(4) represents the increased chat history - with 4 interactions.}
\label{tab:resampling_qwen_gemma}
\end{table}

\section{Transition heatmap}\label{app:unk_to_problem_setup}
\begin{figure}[htbp]
    \centering
    \includegraphics[width=0.8\textwidth]{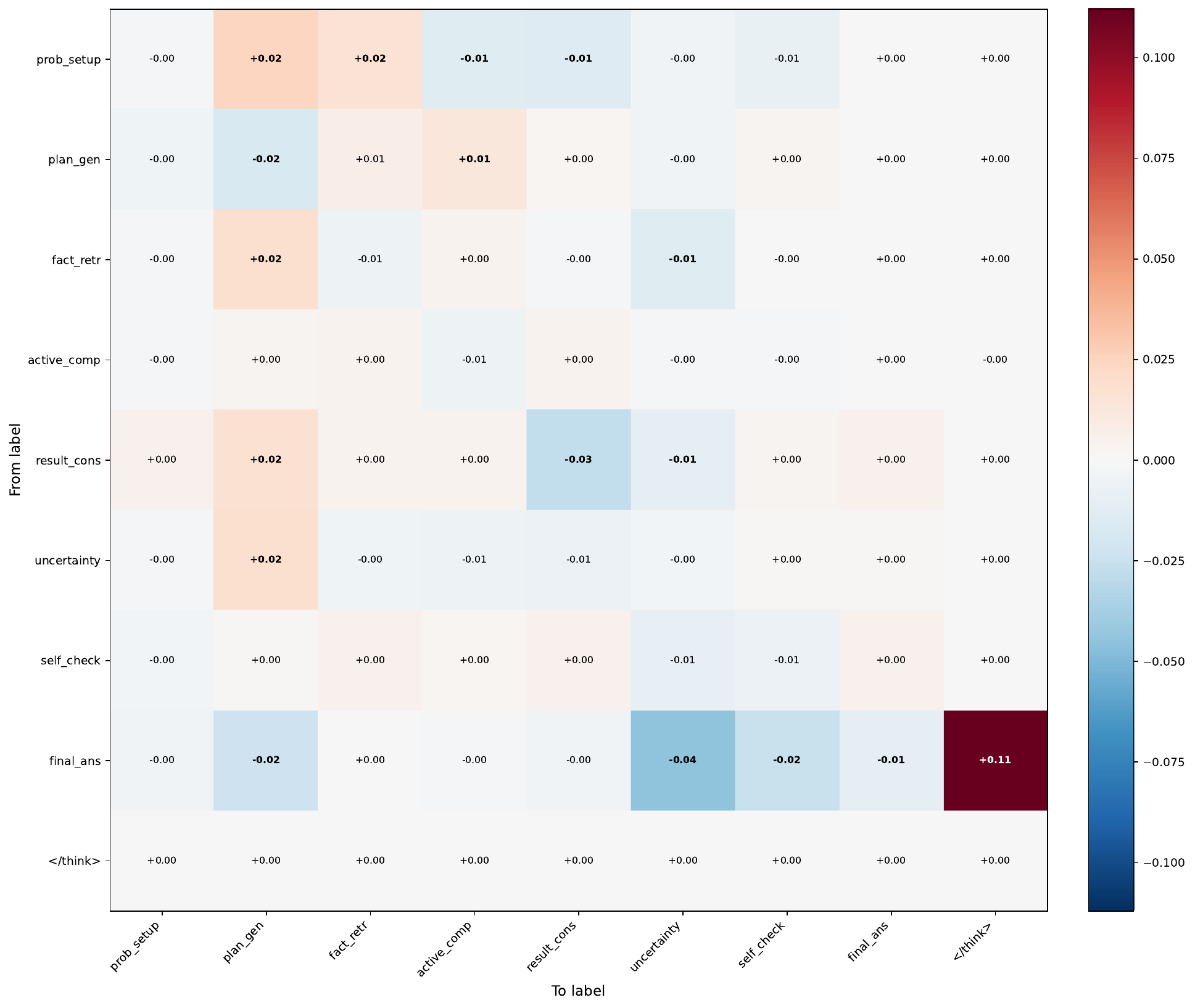}
    \caption{Difference of transition probability matrices (Long Input - Baseline). Qwen3-32B, MATH-500 problems.}
    \label{fig:heatmap}
\end{figure}

Figure \ref{fig:heatmap} represents the difference of transition probability matrices. Positive values mean that Long Input setup has a higher probability of this transition than the Baseline. For this heatmap, we evaluate 500 traces for each context condition, which gives 110,867 transitions for Baseline and 80,508 transitions for Long Input. We evaluate a statistical significance of increased probability of transition from final answer emission to the end of thinking trace as $p = 0.0011$ using Mann-Whitney U test (one-sided (Long Input > Baseline)).

\section{Long Input references}\label{app:long_input_refs}

In this section, we provide all references to the inserted prompt tokens in the \textbf{Long Input} setup made by Qwen3-32B on the MATH-500 benchmark. We note that other models may exhibit different patterns of prompt analysis. For example, Qwen3.5-27B tends to first outline the overall structure of the prompt before beginning to solve the task, typically identifying the distractor tokens early on and concluding that they are irrelevant to the user's request.

\begin{tcolorbox}[colback=gray!5, colframe=gray!75, title=\textbf{Task 203}, sharp corners, breakable]

    \begin{quote}
    <...>
    \newline
    \newline
    Therefore, since f(-2) = 12, the remainder is 12. Therefore, the answer is 12.
    \newline
    \newline
    But let me think if there’s any trick here. The problem is from Shakespeare’s \textit{Coriolanus}? Wait, no, the original problem is just a math problem. The “old data” was the play’s text, but the new task is a math problem. Maybe that was just to set the context? Anyway, regardless, the math seems straightforward.
    \newline
    \newline
    Alternatively, perhaps there is a mistake in assuming that the exponents can be evaluated directly? <...>
    \end{quote}

\end{tcolorbox}

\begin{tcolorbox}[colback=gray!5, colframe=gray!75, title=\textbf{Task 264}, sharp corners, breakable]

    \begin{quote}
    <...>
    \newline
    \newline
    This suggests that the construction in the Asymptote code might have an error, or "bisectorpoint" is not what I think it is. Given the time I've spent and the complexity, I might need to switch strategies.
    \newline
    \newline
    Let me recall that in the problem, after the battle, Coriolanus is made consul, and the citizens are angry, leading to his exile, but this is unrelated to the geometry problem. Focus on the geometry.
    \newline
    \newline
    Given the square CDEG with CD = 3, and rectangle BEFH with BE = 5, find BH. <...>
    \end{quote}

\end{tcolorbox}

\begin{tcolorbox}[colback=gray!5, colframe=gray!75, title=\textbf{Task 296}, sharp corners, breakable]

    \begin{quote}
    <...>
    \newline
    \newline
    Therefore, equation B has the largest x. But let me check another angle.
    \newline
    \newline
    Wait, the problem is from the SAT or similar? Wait, the original problem is from the Coriolanus play? No, the original data is the play, but the new task is a math problem. Anyway, according to my reasoning, equation B should be the answer. But let me check once more.
    \newline
    \newline
    Wait, equation E: $base_E = 1 + 1/r$. For r approaching 0, this goes to infinity <...>
    \end{quote}

\end{tcolorbox}

\begin{tcolorbox}[colback=gray!5, colframe=gray!75, title=\textbf{Task 420}, sharp corners, breakable]

    \begin{quote}
    
    <...> Alternatively, given that the answer requires boxed, maybe they want the scalar? Let me check the problem again.
    \newline
    \newline
    Wait, in the initial problem statement, the user provided a lot of text from Shakespeare's Coriolanus, but the actual math problem is separate. So maybe the problem is from a math textbook, and the user just included some old data. Anyway, focusing on the math problem.
    \newline
    \newline
    Given that the problem says "projection of a onto b," and given that the dot product is given <...>
    \end{quote}

\end{tcolorbox}

\section{Examples of resampled traces}\label{app:resampling_samples}

In this section, we provide several examples of resampled traces, described in Section \ref{sect:analysis}.

\begin{tcolorbox}[colback=gray!5, colframe=gray!75, title=\textbf{Task 3}, sharp corners, breakable]

    \begin{quote}
    \textbf{Baseline:}
    \newline
    <...> That's 9 combinations, which correspond to the 9 divisors. So that's another way to see it. So that confirms it again.
    \newline
    \newline
    Therefore, I think the answer is 9. \textcolor{red!100}{But let me just check once more.} <...>
    \newline
    \newline
    \textbf{Long Input (128 tokens inserted):}
    \newline
    <...> That's 3*3=9 divisors, which again gives me 9. So that's consistent. Therefore, I think 9 is the right answer.
    \newline
    \newline
    \textcolor{red!100}{But just to be thorough, maybe I can check another way.} <...>
    \newline
    \newline
    \textbf{Long Input (16k tokens inserted):}
    \newline
    <...> That's nine combinations, so nine divisors. Yep, that checks out. So I think the answer is 9.
    \newline
    \newline
    **Final Answer**
    The number of positive whole-number divisors of 196 is \boxed{9}.
     \textbf{</think>}
    \end{quote}
    
\end{tcolorbox}

\begin{tcolorbox}[colback=gray!5, colframe=gray!75, title=\textbf{Task 6}, sharp corners, breakable]

    \begin{quote}
    \textbf{Baseline:}
    \newline
    <...> multiples of 3, 27 must be the smallest. 
    \newline
    \newline
    \textcolor{red!100}{But let me think again. Let me check for example negative integers. } <...>
    \newline
    \newline
    \textbf{Long Input (128 tokens inserted):}
    \newline
    <...> multiples of 3, 27 is the smallest. 
    \newline
    \newline
    \textcolor{red!100}{But wait, just to make sure, let me check if maybe 0 is considered.} <...>
    \newline
    \newline
    \textbf{Long Input (16k tokens inserted):}
    \newline
    <...> multiples of 3, 27 is the smallest. 
    \newline
    \newline
    Therefore, the answer is 27.
    \newline
    \newline
    **Final Answer**
    \newline
    The smallest positive perfect cube that can be written as the sum of three consecutive  integers is \boxed{27}.
    \textbf{</think>}

    \end{quote}
    
\end{tcolorbox}

\section{Additional details on fine-tuning}\label{app:fine_tuning}

We base our experiments on the \textit{open-instruct} and \textit{OLMo-core} repositories \citep{olmo3} (version \texttt{v2.5.0}).

\textbf{Training.} In total, we fine-tune \texttt{Olmo-3-7B-Think-SFT} on an augmented 10\% subset of the original \texttt{allenai/Dolci-Think-SFT-7B} training dataset for 7864 optimization steps, corresponding to 8.24B training tokens and two epochs, i.e., the same number of SFT epochs used in the original training of \texttt{Olmo-3-7B-Think-SFT}.

We fine-tune the model using a reduced learning rate of $10^{-6}$, which is lower than the learning rates commonly used in OLMo-core SFT training. We intentionally use a conservative optimization setup to reduce the risk of drifting too far from the original checkpoint and degrading the model's overall reasoning quality. Following the recommendations provided in the OLMo-core repository, we set \texttt{global\_batch\_size} to 1048576 and \texttt{seq\_len} to 32768. Training required approximately one day on 8$\times$H100 GPUs.

\textbf{Evaluation.} 
To verify that the additional training does not degrade the model’s general reasoning capabilities, we  evaluate the resulting model on the GPQA-Diamond benchmark, which consists of 198 PhD-level multiple-choice questions, as well as on MMLU-Pro \citep{wang2024mmlu} using 500 randomly sampled examples. We use the \texttt{Idavidrein/gpqa} and \texttt{TIGER-Lab/MMLU-Pro} benchmark versions from Hugging Face. For MMLU-Pro, the mean number of reasoning tokens is computed only over samples whose total sequence length is shorter than 32k tokens.

Table~\ref{tab:multiturn_sft_gpqa} shows that the proposed training procedure preserves both performance and reasoning length on GPQA-Diamond and MMLU-Pro. Additionally, the Baseline column in Table~\ref{tab:multiturn_sft_math500} demonstrates that performance on MATH-500 is preserved.

\begin{table}[t]
\centering

\small
\begin{tabular}{@{}lcccc@{}}
\toprule
\textbf{Model} 
& \multicolumn{2}{c}{\textbf{GPQA-Diamond}}
& \multicolumn{2}{c}{\textbf{MMLU-Pro}} \\
\cmidrule(lr){2-3}
\cmidrule(lr){4-5}
& \textbf{Acc.} & \textbf{Tok.}
& \textbf{Acc.} & \textbf{Tok.} \\
\midrule

Olmo-3-7B-Think-SFT 
& 42.6
& 9,222
& 49.2
& 1779 \\

Olmo-3-7B-Think-SFT-Ours
& 43.1
& 9,164
& 54.2
& 1912 \\

\bottomrule
\end{tabular}
\caption{GPQA-Diamond and MMLU-Pro performance of the original Olmo-3-7B-Think-SFT model and the fine-tuned variant. Tokens denote the average number of generated reasoning tokens.}
\label{tab:multiturn_sft_gpqa}
\end{table}

\newpage
\section{Task-wise comparison of reasoning lengths}\label{app:task_wise_plots}
\begin{figure}[h]
    \centering

    \begin{subfigure}[b]{0.45\textwidth}
        \centering
        \includegraphics[width=\textwidth]{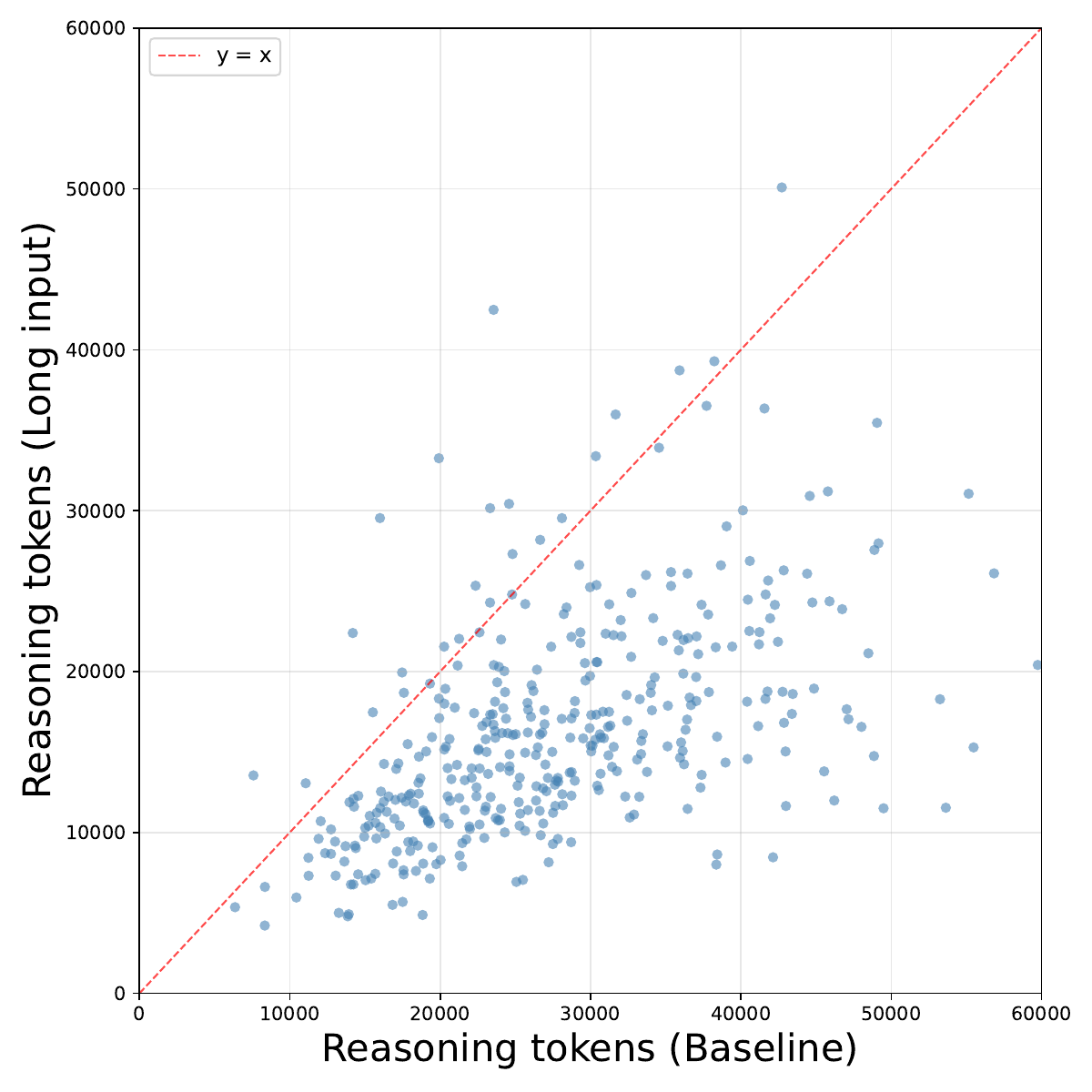}
        \caption{Qwen3.5-27B}
        \label{fig:sub1}
    \end{subfigure}
    \hfill 
    \begin{subfigure}[b]{0.45\textwidth}
        \centering
        \includegraphics[width=\textwidth]{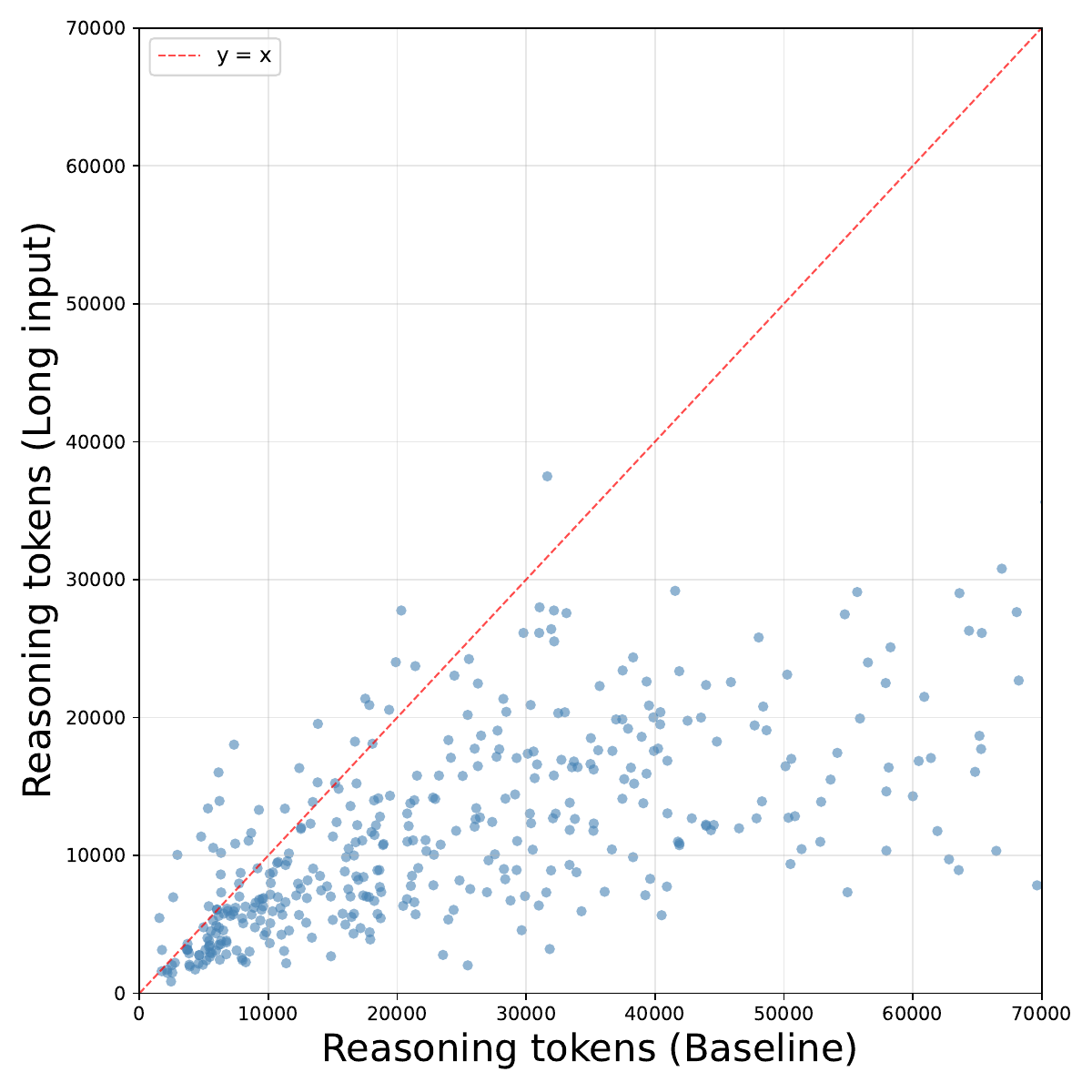}
        \caption{GPT-OSS-120B}
        \label{fig:sub2}
    \end{subfigure}

    \medskip 

    \begin{subfigure}[b]{0.45\textwidth}
        \centering
        \includegraphics[width=\textwidth]{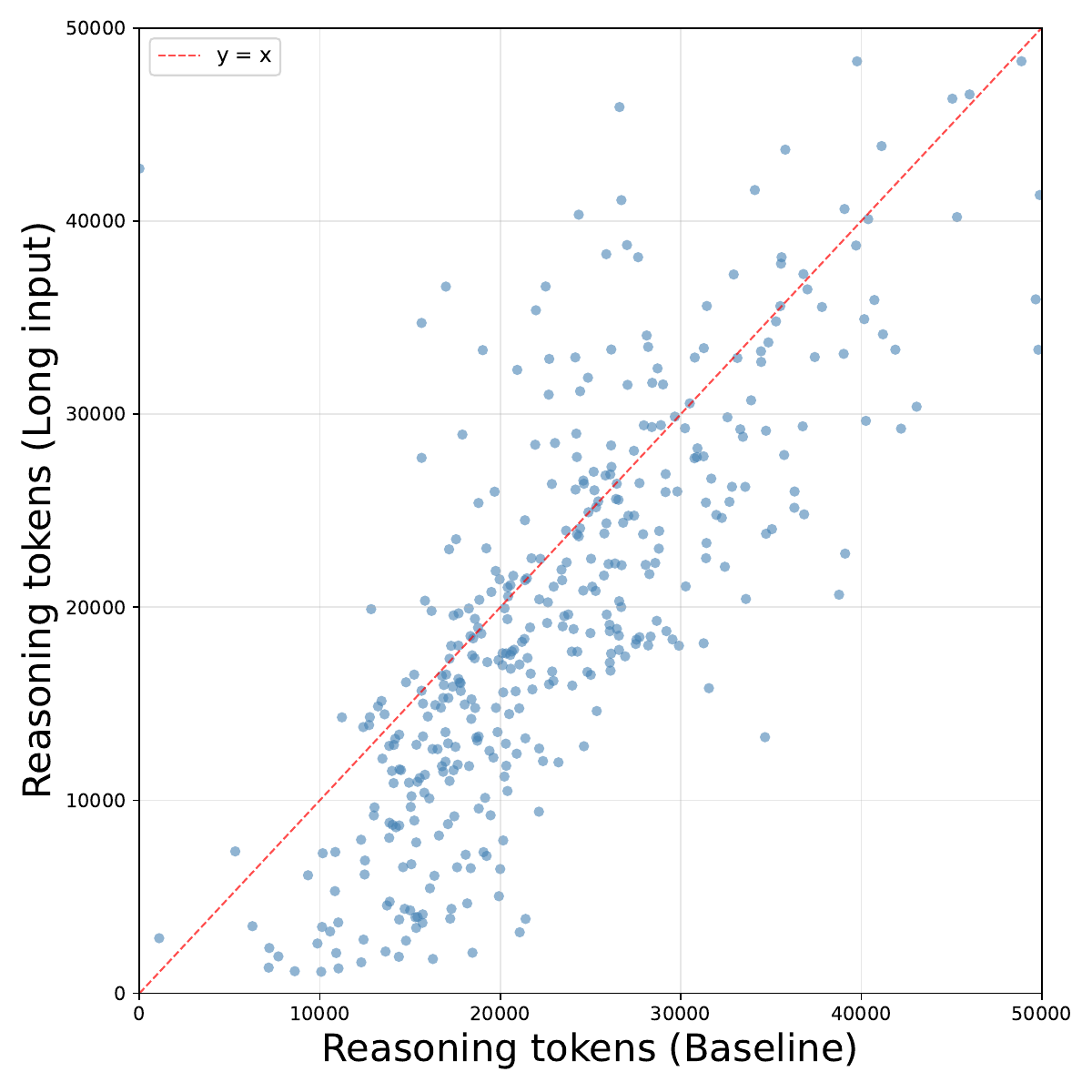}
        \caption{Gemini 3 Flash Preview}
        \label{fig:sub3}
    \end{subfigure}
    \hfill
    \begin{subfigure}[b]{0.45\textwidth}
        \centering
        \includegraphics[width=\textwidth]{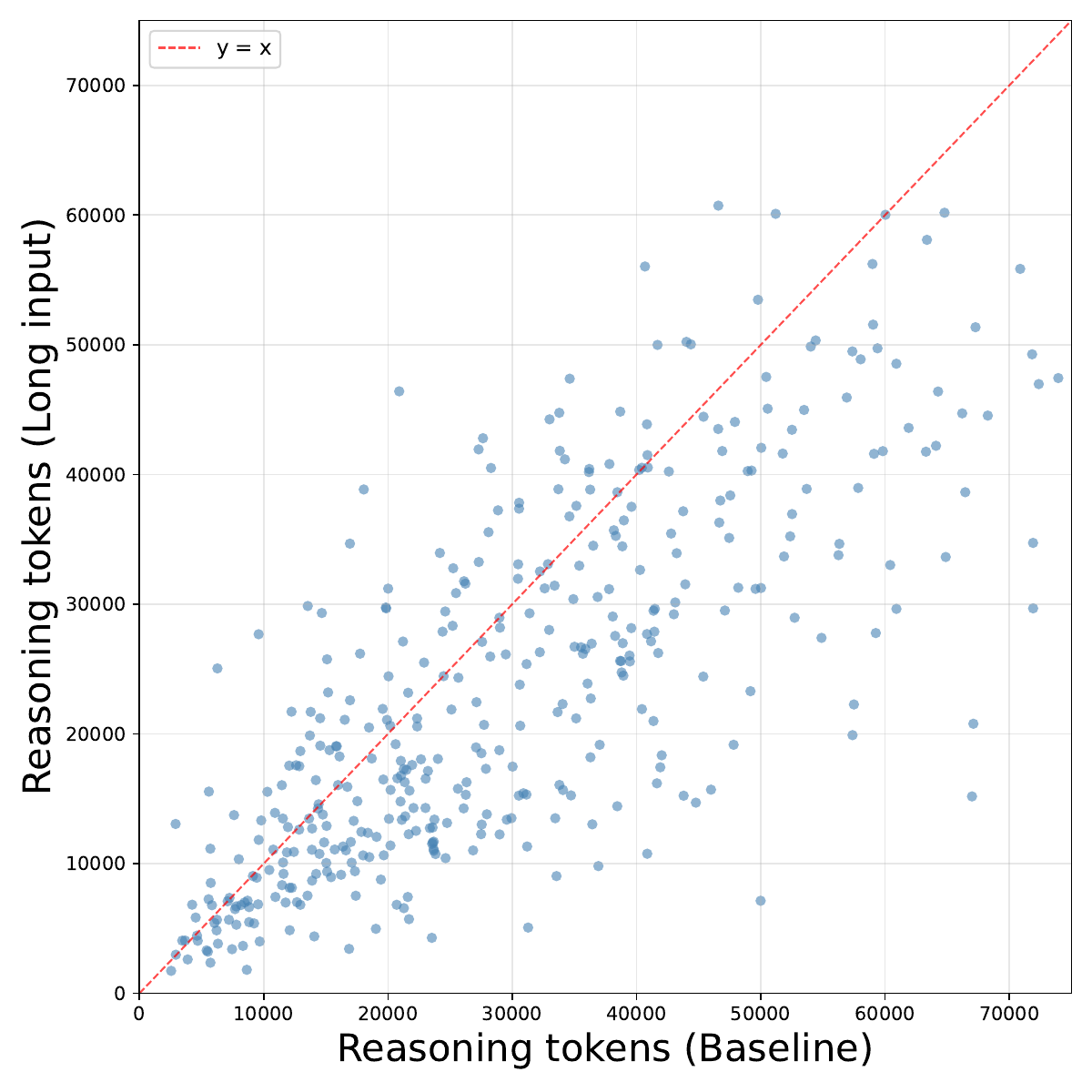}
        \caption{Kimi K2 Thinking}
        \label{fig:sub4}
    \end{subfigure}

    \caption{Number of generated tokens for each IMOAnswerBench task. X-axis: Baseline, Y-Axis: Long Input.}
    \label{fig:main_exp_task_wise}
\end{figure}


\end{document}